\documentclass[9pt]{osa-supplemental-document}
\setboolean{shortarticle}{false}
\usepackage{tikz}
\usetikzlibrary{arrows.meta,positioning,fit,shapes.geometric,calc}
\usepackage{tabularx}

\title{On-Orbit Space AI: Federated, Multi-Agent, and Collaborative Algorithms for Satellite Constellations}
\author{Ziyang Wang} 

\begin{abstract}
Satellite constellations are transforming space systems from isolated spacecraft into networked, software-defined platforms capable of on-orbit perception, decision making, and adaptation. Yet much of the existing AI studies remains centered on single-satellite inference, while constellation-scale autonomy introduces fundamentally new algorithmic requirements: learning and coordination under dynamic inter-satellite connectivity, strict SWaP-C limits, radiation-induced faults, non-IID data, concept drift, and safety-critical operational constraints. This survey consolidates the emerging field of on-orbit space AI through three complementary paradigms: (i) {federated learning} for cross-satellite training, personalization, and secure aggregation; (ii) {multi-agent algorithms} for cooperative planning, resource allocation, scheduling, formation control, and collision avoidance; and (iii) {collaborative sensing and distributed inference} for multi-satellite fusion, tracking, split/early-exit inference, and cross-layer co-design with constellation networking. We provide a system-level view and a taxonomy that unifies collaboration architectures, temporal mechanisms, and trust models. To support community development and keep this review actionable over time, we continuously curate relevant papers and resources at \url{https://github.com/ziyangwang007/AI4Space}.
\end{abstract}

\setboolean{displaycopyright}{false} 

\begin{document}

\maketitle
\section{Introduction}

\begin{figure}
    \centering
    \includegraphics[width=\linewidth]{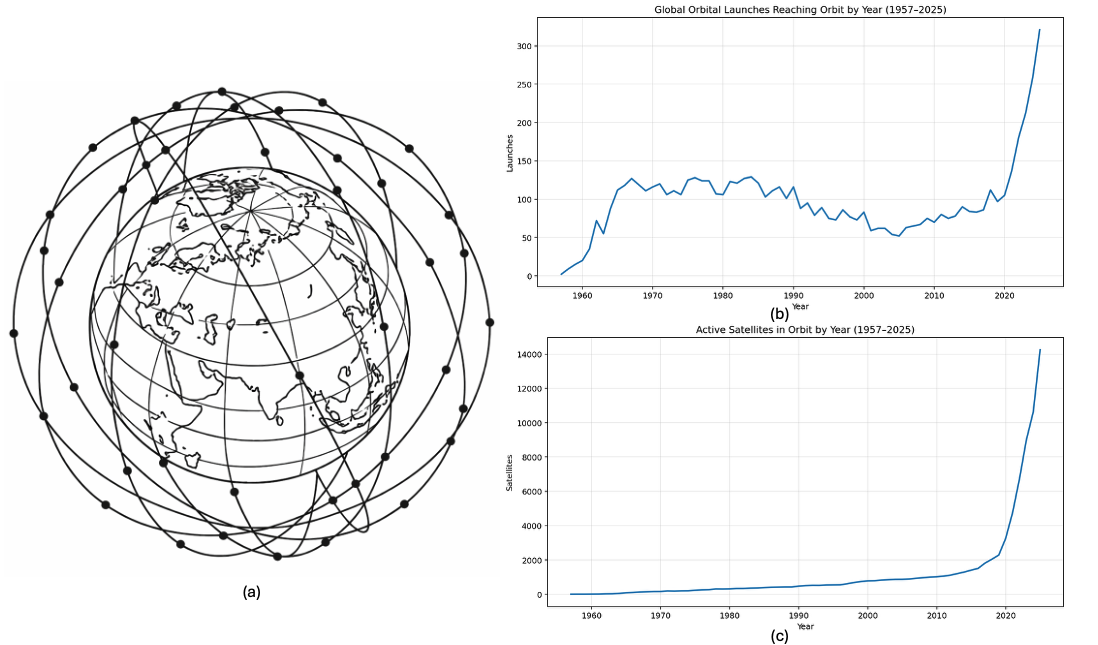}
    \caption{The Brief Introduction of Satellite Constellations. (a) Satellite Constellations, (b) Global Orbital Launches Reaching Orbit (1957-2025), (c) Active Satellites in Orbit (1957-2025).  }
    \label{fig:intro}
\end{figure}

Satellite constellations are increasingly operated as networked, software-defined space systems rather than isolated spacecraft, enabled by platform miniaturization, inter-satellite links (ISLs), and programmable space networking and payload architectures \cite{curzi2020large,kulu2024constellations,radhakrishnan2016isc,chen2024isls,yuan2022sistn}. In parallel, on-orbit/edge computing and onboard AI have moved from conceptual studies to flight demonstrations, making it practical to execute parts of the perception--analysis--action loop in orbit (e.g., cloud masking, event detection, and product prioritization) to reduce downlink volume and shorten the response cycle for time-critical applications \cite{esa_phisat,esa_phisats_programme,giuffrida2022phisat1,esa_opssat,chintalapati2025onboard,yin2025oec,wu2023oec}. Practical flight demonstrations already illustrate this trend at the single-spacecraft level. For example, onboard cloud screening and data prioritization have been demonstrated in orbit, while in-orbit experimentation platforms have provided a testbed for autonomous onboard software; together, these efforts motivate the next step from single-satellite intelligence toward constellation-level distributed collaboration. A brief illustration of satellite constellations\footnote{Source: Our World in Data, Jonathan’s Space Report.} is illustrated in Figure \ref{fig:intro}.

Within the broader theme of Space AI \cite{wang2025space}, these developments raise a central algorithmic question: how can multiple satellites {learn, infer, and decide collaboratively} under time-varying connectivity, strict SWaP-C budgets, radiation-induced faults, non-IID and drifting data streams, and safety-/security-critical operational constraints \cite{nasa2024ai_space,souvatzoglou2024seu,matthiesen2024satfl,zou2024noniid,tran2026conceptdrift}. This survey focuses on on-orbit space AI, i.e., methods that treat a constellation as a distributed system for learning, inference, and autonomy. Compared with single-satellite onboard intelligence, constellation-scale methods must explicitly reason about dynamic contact graphs and intermittent/long-delay links \cite{radhakrishnan2016isc,burleigh2003dtn}, heterogeneous sensing/compute/energy capabilities, and stronger trust requirements: systems should be robust to benign faults and, where relevant, to adversarial or compromised participants while supporting controlled, auditable model updates \cite{bonawitz2017secureagg,blanchard2017krum}.

\subsection{Motivation and Scope}

We consider distributed satellite systems and constellations in which each spacecraft can sense, compute, and communicate with neighboring satellites via inter-satellite links (ISLs) and/or with ground stations \cite{radhakrishnan2016isc,burleigh2003dtn,golkar2015fss,diana2024onboardnn}. Collaboration is valuable because many core mission functions are inherently distributed: (i) {coverage, tasking, and observation scheduling} across many platforms \cite{picard2022eoscsp,yang2023constellation_mission}; (ii) {cooperative observation, fusion, and tracking} where multi-sensor or multi-view aggregation improves estimation quality and robustness \cite{wei2018debris,jia2014fusion,fantacci2015rfs}; and (iii) {resilient networking and resource sharing} where routing, contact utilization, and opportunistic coalitions determine end-to-end latency and throughput \cite{ruizdeazua2021federation,cen2024satflow}. These capabilities are also tightly coupled with safety- and operations-critical constraints (e.g., maneuver conflicts, risk management, and supervisory autonomy), motivating coordination mechanisms that remain effective under partial observability and intermittent connectivity \cite{hilton2024mission,thangavel2024taso}.

The scope of this review is algorithmic. We emphasize methods intended to run {on orbit}, either fully onboard or jointly with the ground segment, but with the constellation itself playing an active role in computation and coordination \cite{wu2023oec}. We organize the review around three paradigms. The first is {federated learning} for cross-satellite training and adaptation without centralizing raw data \cite{mcmahan2017fedavg,matthiesen2024satfl}. The second is {multi-agent algorithms} for cooperative autonomy, planning, scheduling, and resource allocation under constraints \cite{hernandezleal2019marl_survey,choi2009cbba,picard2022eoscsp}. The third is {collaborative sensing and distributed inference} for multi-satellite fusion, tracking, and partitioned/split inference \cite{wei2018debris,fantacci2015rfs,branchynet2017,spaceexit2025}. We do not attempt to comprehensively survey single-satellite onboard vision models, purely hardware-focused studies, or ground-only analytics, except when they directly affect the design and evaluation of distributed on-orbit algorithms \cite{ ,chintalapati2025onboard}.

\subsection{Key Definitions: Federated vs Multi-Agent vs Collaborative}
To keep terminology consistent, we use the following definitions.

Distributed on-orbit space AI refers to methods in which multiple satellites (optionally with ground nodes) jointly perform learning, inference, and/or decision-making, and where computation and communication are distributed under orbital contact graphs, DTN-like intermittency, and space operational constraints \cite{golkar2015fss,radhakrishnan2016isc,burleigh2003dtn,thangavel2024taso}.

Federated learning (FL) refers to collaborative training or adaptation where each satellite keeps its raw data locally and exchanges model-related information (e.g., parameters, gradients, updates, or distilled knowledge). Aggregation may happen on the ground, through a hierarchical structure (e.g., cluster-head satellites), or via decentralized protocols \cite{mcmahan2017fedavg,matthiesen2024satfl}. In constellation settings, FL must address intermittent participation, asynchronous communication, non-IID data, and safe/secure update mechanisms (e.g., secure aggregation and robust aggregation under unreliable participants) \cite{bonawitz2017secureagg}.

Multi-agent algorithms model each satellite as an agent that selects actions (e.g., sensing schedules, pointing commands, routing decisions, maneuvering plans, or compute allocations) to optimize mission objectives under constraints. Cooperation can be achieved through multi-agent reinforcement learning, distributed planning, auction/market-based task allocation, or distributed control \cite{hernandezleal2019marl_survey,gronauer2022marl_survey,picard2022eoscsp,yang2023constellation_mission}. Communication is usually limited, delayed, and topology-dependent, so when and what to share becomes part of the algorithm design.

Collaborative sensing and distributed inference focus on sharing and fusing information for perception and estimation, and on distributing inference computation across nodes. Examples include distributed filtering and tracking, multi-view fusion, feature-level exchange, and split/early-exit inference \cite{wei2018debris,jia2014fusion,fantacci2015rfs,branchynet2017,spaceexit2025}. This paradigm often requires joint design with networking and orbital edge computing, because communication choices directly affect inference quality, latency, and energy \cite{yin2025oec,cen2024satflow}.

\begin{figure}
    \centering
    \includegraphics[width=\linewidth]{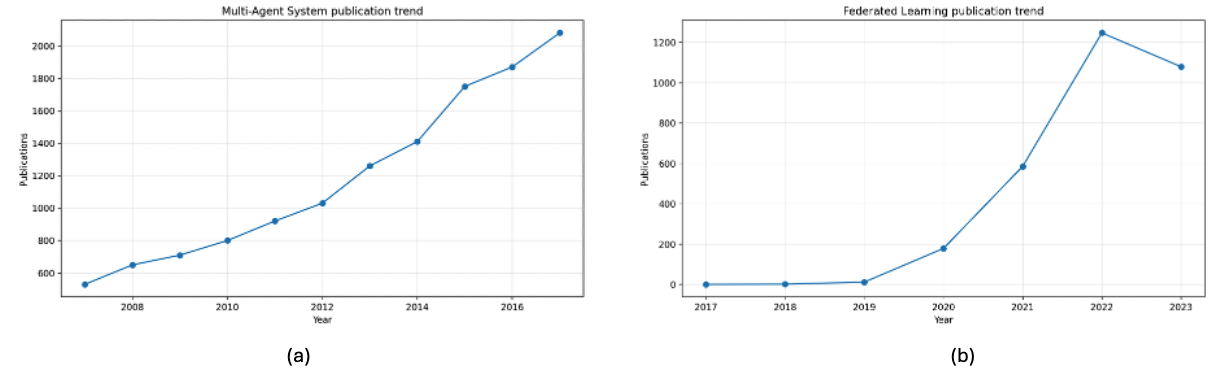}
    \caption{The Publication Trend of AI. (a) Federated learning, (b) Multi-agent system. }
    \label{fig:num}
\end{figure}

These paradigms are not mutually exclusive. Federated learning can train perception models used by multi-agent autonomy; multi-agent coordination can determine which satellites participate in learning; and collaborative inference can be combined with federated updates to sustain performance under drift and changing contact patterns \cite{thangavel2024taso,yin2025oec}. A brief line chart of the publication number based on Web of Science database around Federated learning and Multi-agent system is illustrated in Figure \ref{fig:num}.

\begin{table*}[htbp]
\centering
\caption{Qualitative Comparison of AI Architectures for Orbital Space Systems}
\label{tab:orbital_ai_comparison}
\small
\setlength{\tabcolsep}{4pt}
\renewcommand{\arraystretch}{1.2}
\begin{tabularx}{\textwidth}{>{\raggedright\arraybackslash}p{2.9cm}
>{\raggedright\arraybackslash}X
>{\raggedright\arraybackslash}X
>{\raggedright\arraybackslash}X
>{\raggedright\arraybackslash}X
>{\raggedright\arraybackslash}X
>{\raggedright\arraybackslash}X}
\toprule
\textbf{Property} 
& \textbf{Ground-centric centralized AI} 
& \textbf{Single-satellite onboard AI} 
& \textbf{Federated learning} 
& \textbf{Collaborative sensing} 
& \textbf{Distributed inference} 
& \textbf{Multi-agent autonomous constellation} \\
\midrule
Communication burden 
& Very high 
& Low 
& Moderate 
& High 
& Moderate 
& Moderate \\

Inter-satellite link dependence 
& Low 
& Low 
& Moderate 
& Moderate 
& High 
& High \\

End-to-end latency 
& Very high 
& Low 
& Moderate 
& Moderate 
& Low 
& Low \\

Privacy / data sovereignty 
& Very low 
& High 
& High 
& Moderate 
& High 
& High \\

Onboard compute \& energy pressure 
& Low 
& Very high 
& High 
& Moderate 
& Moderate 
& Moderate \\

Scalability to large constellations 
& Moderate 
& Moderate 
& High 
& High 
& High 
& High \\

Fault tolerance / resilience 
& Low 
& Moderate 
& Moderate--High 
& High 
& High 
& Very high \\

Deployment \& verification complexity 
& Low 
& Moderate 
& High 
& High 
& High 
& Very high \\

Autonomy level 
& Very low 
& High 
& Moderate 
& Moderate 
& High 
& Very high \\

Current feasibility / maturity 
& Very high 
& High 
& Moderate 
& Moderate 
& Moderate 
& Moderate \\
\bottomrule
\end{tabularx}
\end{table*}

\subsection{Related Surveys \& Our Positioning}
A growing body of surveys has examined adjacent aspects of ``AI + satellite constellations,'' but often from a single-layer perspective. 
In satellite networking and non-terrestrial networks (NTN), surveys largely focus on AI/ML as a tool for communications, routing, resource management, and integration with terrestrial systems \cite{liu2018sagin,mahboob2024ntn_survey,fontanesi2025ai_satcom_survey,chen2023sagin_survey,diana2024onboardnn}. 
In orbital/space edge computing, the emphasis is typically on system architectures and computation offloading/scheduling under time-varying connectivity \cite{yin2025oec,wu2023oec,shen2023sagin_computing}. 
A separate line of work reviews onboard AI from a device/software viewpoint (accelerators, compression, deployment constraints), mostly at the single-satellite level \cite{ ,duggan2025ejrs}. 
Finally, constellation-specific federated learning surveys concentrate on distributed training protocols under intermittent contacts and non-IID data, but do not aim to cover constellation-scale decision-making and collaborative inference as a unified algorithmic stack \cite{matthiesen2024satfl}.

Table~\ref{tab:related_surveys} summarizes representative related surveys and highlights the gap this paper addresses.
In contrast to prior work, we take a {constellation-as-a-distributed-system} view and unify three algorithmic pillars---\emph{federated learning}, {multi-agent coordination}, and {collaborative sensing/distributed inference}---under a shared taxonomy of {architecture} (centralized/hierarchical/decentralized), {timing} (synchronous/asynchronous/event-driven), and {trust} (benign faults to Byzantine/adversarial). 
This positioning is intended to make methods comparable across communities (distributed optimization, multi-agent systems, and distributed estimation/inference), and to provide evaluation guidance that reflects realistic on-orbit constraints.

\begin{table}[H]
\caption{Comparison with representative related surveys. \textbf{Y}: explicit focus; \textbf{P}: partial/adjacent coverage; \textbf{--}: not a focus.}
\label{tab:related_surveys}
\centering
\footnotesize
\resizebox{\textwidth}{!}{%
\begin{tabular}{l c p{4.0cm} c c c c c c}
\hline
\textbf{Survey} & \textbf{Year} & \textbf{Primary scope} 
& \textbf{On-orbit/Edge} & \textbf{FL} & \textbf{Multi-agent} & \textbf{Collab. Inf./Fusion} & \textbf{Trust/Safety} & \textbf{Bench./Twin} \\
\hline
Liu et al. \cite{liu2018sagin} & 2018 & SAGIN architecture, resource mgmt., integration & P & -- & -- & -- & P & -- \\
Shen et al. \cite{shen2023sagin_computing} & 2023 & Computing tech in SAGIN (cloud/edge synergy, enabling tools) & Y & P & -- & -- & P & P \\
Mahboob \& Liu \cite{mahboob2024ntn_survey} & 2024 & AI-enabled satellite-based NTN for 6G (networking-centric) & P & P & -- & -- & P & -- \\
Fontanesi et al. \cite{fontanesi2025ai_satcom_survey} & 2025 & AI/ML for SATCOM (use cases, architectures, hardware notes) & P & -- & -- & -- & P & -- \\
Yin et al. \cite{yin2025oec} (Wu et al. \cite{wu2023oec}) & 2025 (2023) & Orbital edge computing: systems, offloading, scheduling & Y & P & P & P & P & P \\
Matthiesen et al. \cite{matthiesen2024satfl} & 2024 & Federated learning in satellite constellations & P & Y & -- & -- & P & P \\
Duggan et al. \cite{duggan2025ejrs} & 2025 & AI-powered onboard EO image processing (deployment constraints) & Y & -- & -- & P & P & P \\
Thangavel et al. \cite{thangavel2024taso} & 2024 & Trusted autonomous satellite operations (DSS/TASO perspective) & P & -- & P & P & Y & P \\
\hline
\textbf{This survey (ours)} & 2026 & Distributed on-orbit AI for constellations: learning + decision + inference & \textbf{Y} & \textbf{Y} & \textbf{Y} & \textbf{Y} & \textbf{Y} & \textbf{Y} \\
\hline
\end{tabular}%
}
\end{table}

\subsection{Contributions and Paper Organization}

This survey makes the following contributions:
\begin{itemize}
    \item \textbf{System-oriented problem framing.} We provide a constellation-centric view of distributed on-orbit AI, highlighting how orbital motion, intermittent contacts, SWaP-C limits, radiation-related faults, and safety constraints shape algorithm design.
    \item \textbf{Unified taxonomy across learning, decision, and inference.} We introduce a taxonomy and shared formulations that connect (i) distributed optimization for learning, (ii) multi-agent decision-making for coordination, and (iii) distributed inference for collaborative sensing, and categorize methods by {architecture}, {timing}, and {trust} assumptions.
    \item \textbf{Structured synthesis of three paradigms.} We review representative methods in federated and privacy-preserving learning, multi-agent coordination, and collaborative sensing/distributed inference, emphasizing constellation-specific issues such as non-IID data, drift, bandwidth/energy budgets, faults, and safe update governance.
\end{itemize}

The remainder of this paper is organized as follows. Section~2 describes the constellation system model and presents a taxonomy of distributed on-orbit AI settings. Section~3 reviews federated and privacy-preserving learning for satellite constellations. Section~4 covers multi-agent coordination algorithms for constellation autonomy. Section~5 discusses collaborative sensing and distributed inference. Section~6 discussed the open challenges and future directions for academics and industry, and Section~7 concludes the survey.

\section{System Model and Taxonomy}
This section introduces a generic system model for distributed on-orbit AI in satellite constellations and then proposes a taxonomy to organize the design space. The goal is to make later algorithm sections comparable under shared assumptions about networking, onboard resources, and data properties.

\begin{table}[H]
\caption{Notation used in the system model (Section~2).}
\label{tab:notation}
\centering
\footnotesize
\resizebox{\textwidth}{!}{%
\begin{tabular}{p{2.2cm} p{9.6cm} p{3.2cm}}
\hline
\textbf{Symbol} & \textbf{Meaning} & \textbf{Where used} \\
\hline
$\mathcal{S}$ & Set of satellites, $\{1,\dots,N\}$ & Sec.~2.1 \\
$\mathcal{G}$ & Set of ground nodes/stations, $\{1,\dots,M\}$ (note: $\mathcal{G}(t)$ also denotes the time-varying graph) & Sec.~2.1 \\
$\mathcal{V}$ & Node set of the communication graph, $\mathcal{V}=\mathcal{S}\cup\mathcal{G}$ & Sec.~2.1 \\
$\mathcal{E}(t)$ & Edge set at time $t$ (available links) & Sec.~2.1 \\
$\mathcal{G}(t)$ & Time-varying communication graph $(\mathcal{V},\mathcal{E}(t))$ & Sec.~2.1 \\
$B_{ij}(t)$ & Link capacity between nodes $i$ and $j$ at time $t$ & Sec.~2.1 \\
$D_{ij}(t)$ & Link latency between nodes $i$ and $j$ at time $t$ & Sec.~2.1 \\
$P_{ij}(t)$ & Link reliability (e.g., packet loss probability) & Sec.~2.1 \\
$\mathcal{C}$ & Contact plan/schedule (contact windows) & Sec.~2.1 \\
$\mathcal{T}$ & Task stream/set for constellation operations & Sec.~2.1 \\
$k=(w_k,r_k,u_k)$ & Task descriptor: window/constraints, resource cost, utility & Sec.~2.1 \\
$C_i(t)$ & Compute budget of satellite $i$ at time $t$ & Sec.~2.1 \\
$E_i(t)$ & Energy budget of satellite $i$ at time $t$ & Sec.~2.1 \\
$S_i(t)$ & Storage budget of satellite $i$ at time $t$ & Sec.~2.1 \\
$\mathcal{D}_i(t)$ & Local data stream at satellite $i$ & Sec.~2.2 \\
$p_i(x,y,t)$ & Underlying local data distribution at satellite $i$ & Sec.~2.2 \\
$\theta$ & Model parameters & Sec.~3 \\
$F_i(\theta)$ & Local FL objective at satellite $i$ & Sec.~3.1 \\
$w_i$ & Aggregation weight in FL global objective & Sec.~3.1 \\
$a_{ij}(t)$ & Mixing weights for decentralized FL at time $t$ & Sec.~3.1 \\
\hline
\end{tabular}%
}
\end{table}

\subsection{Constellation Setting: Missions, Nodes, and Communication (ISLs, Dynamic Topology)}
We model a constellation as a set of satellites $\mathcal{S}=\{1,\dots,N\}$ and (optionally) a set of ground nodes $\mathcal{G}=\{1,\dots,M\}$. Each satellite is equipped with one or more payloads (e.g., optical, SAR, infrared, GNSS, AIS, RF sensing), an onboard computing unit, local storage, and communication terminals for space-to-space and space-to-ground links. Inter-satellite links (ISLs) and scheduled space networking make the constellation topology time-dependent and operationally constrained \cite{radhakrishnan2016isc,chen2024isls,handley2018delay}.

\textbf{Mission and task abstraction.}
Constellation operations can be viewed as selecting and executing a stream of tasks $\mathcal{T}$, such as observation requests, onboard processing, cooperative tracking, and communication/routing actions. Following the standard abstraction in satellite scheduling and mission planning, we represent each task $k\in\mathcal{T}$ by a feasible opportunity window and constraints, a resource-cost vector, and a utility value \cite{lemaitre2002agile,ferrari2025ssp}:
\begin{equation}
k = \big(w_k,\, r_k,\, u_k\big),
\end{equation}
where $w_k$ specifies the valid time window (and associated geometric/pointing constraints), $r_k$ summarizes required resources (e.g., compute cycles, energy, storage, and communication volume), and $u_k$ denotes the mission utility or priority. This abstraction provides a common language for later sections on scheduling, resource allocation, and contact-aware learning/inference.

\textbf{Time-varying connectivity and contact windows.}
ISLs induce a time-varying communication graph
\begin{equation}
\mathcal{G}(t) = \big(\mathcal{V},\,\mathcal{E}(t)\big), \quad \mathcal{V}=\mathcal{S}\cup\mathcal{G},
\end{equation}
where an edge $(i,j)\in\mathcal{E}(t)$ exists when a communication opportunity is available between nodes $i$ and $j$ at time $t$ (line-of-sight, pointing constraints, link scheduling). For an active link, we characterize its time-dependent capacity $B_{ij}(t)$, latency $D_{ij}(t)$, and reliability $P_{ij}(t)$; these vary across RF/optical ISLs and relay/direct-to-ground modes \cite{radhakrishnan2016isc,handley2018delay}.

In many constellations, connectivity is intermittent and naturally modeled via a delay-/disruption-tolerant networking (DTN) perspective, where routing and aggregation are planned around contact windows rather than assuming persistent end-to-end paths \cite{burleigh2003dtn,rfc4838}. A common operational abstraction is a {contact plan} (or contact schedule),
\begin{equation}
\mathcal{C}=\Big\{ \big(i,j, t^{\mathrm{start}}, t^{\mathrm{end}}, B_{ij}, D_{ij}\big) \Big\},
\end{equation}
which encodes predictable communication opportunities and enables contact-graph based routing and collaboration scheduling \cite{burleigh2011cgr,araniti2015cgr,birrane2012cgr}. This same contact-plan abstraction is also useful for designing contact-aware learning rounds, hierarchical aggregation, and cooperative inference pipelines in later sections.

\textbf{Constellation geometry and communication patterns.}
Constellation geometry also affects communication strategy. In Walker-Delta constellations, orbital planes are distributed across $360^\circ$, whereas Walker-Star constellations distribute planes across $180^\circ$; as a result, inter-plane connectivity patterns and handover behavior differ between the two. In near-polar Walker-Star systems, cross-seam inter-plane links may connect satellites moving in opposite directions, which increases relative velocity and Doppler and makes persistent inter-plane communication more challenging.

\textbf{Onboard compute, energy, and storage constraints.}
Each satellite $i$ operates under time-varying budgets of compute $C_i(t)$, energy $E_i(t)$, and storage $S_i(t)$, shaped by duty cycling, thermal constraints, payload operations, and link schedules. These constraints directly influence what can be trained or inferred onboard, how often models can be updated, and what intermediate information can be shared \cite{yin2025oec,chintalapati2025onboard}.

\textbf{Where collaboration happens.}
Depending on mission design and governance, collaboration may be primarily (i) space-to-ground (ground-coordinated learning/planning), (ii) space-to-space (ISL-enabled peer/cluster cooperation with limited ground contact), or (iii) hybrid (local autonomy with periodic ground synchronization/validation). The remainder of this survey uses the above graph-and-contact view to compare algorithmic assumptions and feasibility under intermittent connectivity.

\subsection{On-Orbit Constraints and Data Properties (SWaP-C, Radiation, Non-IID, Drift)}
Distributed AI in orbit is governed by coupled constraints across computation, communication, reliability, operations, and data---many of which are weaker or absent in terrestrial settings.

\textbf{SWaP-C limits and communication cost.}
Satellites operate under strict size, weight, power, and cost (SWaP-C) budgets that bound compute throughput, memory, and available energy, and impose strong thermal and duty-cycle constraints on sustained onboard workloads \cite{veyette2022aiml,swope2023issbenchmark,chintalapati2025onboard,duggan2025ejrs}. Communication is similarly expensive and often availability-limited: downlink/ISL sessions are time-windowed and transmission can incur non-trivial power draw during high-rate engagements, which motivates selective sharing (e.g., compact features or model deltas rather than raw data) and event-driven messaging \cite{rose2019oe,yin2025oec}.

\textbf{Intermittent connectivity and long-delay networking.}
Unlike stable terrestrial networks, constellation connectivity is dictated by orbital dynamics and link scheduling, yielding time-varying topologies, long/variable delays, and frequent partitions. Many constellations are therefore well described by delay-/disruption-tolerant networking (DTN) abstractions, where routing and coordination must tolerate intermittent contacts and partial participation \cite{burleigh2003dtn,rfc4838,handley2018delay,araniti2015cgr}. Practically, this pushes distributed learning and coordination toward asynchronous, contact-aware protocols that explicitly handle staleness and missing updates.

\textbf{Radiation-induced faults and unreliable computation.}
The space environment introduces radiation-induced faults (e.g., single-event upsets and latchup-related anomalies) that may manifest as bit flips, crashes, and silent data corruption, even when the network is functioning normally \cite{karnik2004seu,oliveira2016gpu,wang2023mars,souvatzoglou2024seu}. As a result, distributed methods must be robust not only to communication loss but also to transient compute/memory faults and intermittent node resets, motivating update sanity checks, redundancy, and conservative fusion/aggregation.

\textbf{Safety-critical operations and update governance.}
Many constellation decisions are safety- and operations-critical, including collision avoidance, maneuver planning, and actions that affect attitude control, power budgets, and mission safety envelopes. Learning-enabled autonomy must respect hard constraints, support runtime monitoring and conservative fallbacks, and treat model updates as governed, auditable artifacts (e.g., staged rollout and rollback) rather than unconstrained online adaptation \cite{thangavel2024taso,autoca_esa,bourriez2023collision}. In other words, radiation-induced faults mean that abnormal outputs cannot be treated as ordinary statistical noise alone. They must instead be anticipated as a system-design constraint, motivating redundancy, sanity checks, robust aggregation or fusion, and conservative update admission throughout the learning and inference pipeline.

\textbf{Non-IID and drifting data streams.}
Data collected by different satellites are typically non-identically distributed due to orbit geometry (latitude coverage, local time), payload diversity and calibration, operational modes, and regional weather/illumination patterns; distributions also evolve over time. Let $\mathcal{D}_i(t)$ denote the local data stream at satellite $i$ with underlying distribution $p_i(x,y,t)$. Constellations commonly exhibit both cross-satellite heterogeneity and temporal drift:
\begin{equation}
\forall i\neq j:\; p_i(x,y,t)\neq p_j(x,y,t), \qquad \text{and}\qquad \exists\, t\neq t':\; p_i(x,y,t)\neq p_i(x,y,t').
\end{equation}
These properties affect convergence, aggregation bias, and fairness, and motivate personalization and heterogeneity-aware objectives in federated and distributed learning \cite{matthiesen2024satfl,kairouz2021fl}. More broadly, the resulting {concept drift} and domain shift in high-dimensional imagery streams motivate continual adaptation and drift-aware evaluation protocols \cite{tran2026conceptdrift}.

\begin{table}[H]
\caption{Mapping from on-orbit constraints to algorithmic design levers and evaluation signals.}
\label{tab:constraints_map}
\centering
\footnotesize
\resizebox{\textwidth}{!}{%
\begin{tabular}{p{3.2cm} p{6.2cm} p{6.4cm} p{3.0cm}}
\hline
\textbf{Constraint} & \textbf{Algorithmic implications} & \textbf{Typical design levers} & \textbf{Evaluation signals} \\
\hline
SWaP-C (compute/energy/memory) \cite{yin2025oec,chintalapati2025onboard}
& Limits model size, training frequency, and allowable on-orbit optimization; training competes with payload ops and comm sessions
& Lightweight updates (adapters/heads), mixed precision, early-exit, split inference, sparse/quantized updates
& Energy (J/orbit), FLOPs, memory, latency, duty cycle \\

Intermittent connectivity / long delay \cite{rfc4838,handley2018delay}
& No persistent end-to-end paths; partial participation; stale information
& Async protocols, staleness-aware weighting, contact-plan scheduling, DTN-aware routing/aggregation
& Staleness, participation rate, time-to-update, timeliness \\

Limited bandwidth / short contact windows \cite{rfc4838,yin2025oec}
& Update payload size dominates feasibility; large models hard to synchronize
& Quantization/sparsification, low-rank deltas, distillation/logits, feature/tracklet sharing
& Bytes/contact, throughput, accuracy-vs-bytes curve \\

Radiation-induced faults / SDC \cite{karnik2004seu,oliveira2016gpu,souvatzoglou2024seu,wang2023mars}
& Corrupted updates/messages; node resets; unreliable compute
& Sanity checks, redundancy, robust aggregation/fusion, conservative filters (e.g., CI)
& Fault-injection robustness, recovery time, consistency \\

Safety-critical operations \cite{thangavel2024taso,nasa2023ca2}
& Hard constraints must never be violated; updates must be governed
& Safety filters/RTA, staged rollout/rollback, audit logs, conservative fallbacks
& Constraint violations, intervention rate, rollback triggers \\

Non-IID across satellites \cite{matthiesen2024satfl,kairouz2021fl}
& Global model bias; slower convergence; fairness issues
& Personalization, clustering, heterogeneity-aware objectives (FedProx/SCAFFOLD/FedNova)
& Per-node accuracy, fairness, convergence speed \\

Concept drift over time \cite{tran2026conceptdrift}
& Model aging/regression; delayed labels
& Drift-aware evaluation, periodic re-sync, controlled continual learning + validation gates
& Accuracy over time, drift detection delay, stability \\
\hline
\end{tabular}%
}
\end{table}

\subsection{Taxonomy of Distributed On-Orbit AI (Architectures, Timing, Trust)}
To compare methods across federated learning, multi-agent coordination, and collaborative sensing/inference, we organize distributed on-orbit AI along three primary dimensions: {collaboration architecture}, {timing mechanism}, and {trust model}. For practical feasibility comparisons, we also track {what is shared} (data/features/models/decisions), since message semantics and size often dominate SWaP-C and contact-window budgets.

A compact summary is provided in Table~\ref{tab:taxonomy_orbit}. In brief, collaboration architecture describes where orchestration/aggregation happens: {centralized} (typically ground-orchestrated), {hierarchical} (e.g., cluster-head satellites or multi-tier aggregators), and {decentralized} (peer-to-peer consensus/gossip), often in hybrid combinations \cite{matthiesen2024satfl,hisatfl2025,lian2017dpsgd}. Timing mechanisms capture how updates are triggered under intermittent contacts: {synchronous} rounds (simple but fragile), {asynchronous} updates (staleness-aware), {event-triggered} communication/coordination (uncertainty- or value-driven), and {orbit-periodic} schedules aligned with predictable contact patterns \cite{mcmahan2017fedavg,wu2025sflleo,hu2023etcn}. Trust models distinguish benign unreliability from adversarial behavior, ranging from {benign faults} to {honest-but-curious} participants (privacy threats) and {Byzantine/compromised} nodes (poisoning or misinformation), which motivates secure and robust aggregation/fusion \cite{bonawitz2017secureagg,blanchard2017krum}.

\begin{table}[H]
\caption{Taxonomy for distributed on-orbit AI in satellite constellations.}
\label{tab:taxonomy_orbit}
\centering
\footnotesize
\resizebox{\textwidth}{!}{%
\begin{tabular}{p{3.0cm} p{5.2cm} p{7.8cm}}
\hline
\textbf{Dimension} & \textbf{Categories} & \textbf{Implications for design and evaluation} \\
\hline
Architecture &
Centralized; Hierarchical; Decentralized (often hybrid) &
Determines orchestration/aggregation points, failure modes, governance, and communication hotspots; affects latency and resilience under ground outages \cite{matthiesen2024satfl,hisatfl2025,lian2017dpsgd}. \\
\hline
Timing &
Synchronous; Asynchronous; Event-triggered; Orbit-periodic &
Controls tolerance to intermittency and staleness; enables contact-aware scheduling (orbit-periodic) and value-driven messaging (event-triggered) \cite{mcmahan2017fedavg,wu2025sflleo,hu2023etcn}. \\
\hline
Trust &
Benign faults; Honest-but-curious; Byzantine/compromised &
Dictates whether privacy-preserving and robust aggregation/fusion are required; impacts update validation, authentication, and anomaly filtering \cite{bonawitz2017secureagg,blanchard2017krum}. \\
\hline
What is shared &
Raw data; Features/statistics; Model updates; Decisions/intents &
Directly determines message size and feasibility under SWaP-C/contact windows; links to split/early-exit inference, distributed tracking, and coordination messaging \cite{fantacci2015rfs,branchynet2017,spaceexit2025,choi2009cbba}. \\
\hline
\end{tabular}%
}
\end{table}

\subsection{Practical Communication Constraints for On-Orbit Collaboration}

A defining difference between terrestrial distributed AI and constellation-scale on-orbit collaboration is that communication cannot be treated as a transparent background service. In satellite constellations, latency, bandwidth, and connectivity are all strongly shaped by orbital motion, scheduled contacts, and link switching, so communication becomes a first-order algorithmic constraint rather than merely an implementation detail \cite{rfc4838,handley2018delay,chen2024isls,cen2024satflow}. In practice, this means that the performance of federated learning, multi-agent coordination, and collaborative sensing is often limited less by nominal model quality than by when information can be exchanged, how much can be exchanged, and how stale that information becomes before it is used.

\textbf{Latency and staleness.}
Even when communication is available, propagation delay, queueing, store-and-forward routing, and deferred contact opportunities can make shared information stale by the time it is consumed. In federated learning, this appears as stale model updates and delayed aggregation, which can slow convergence or bias the global model toward recently connected participants \cite{razmi2022groundassisted,xie2019async,nguyen2022fedbuff,lin2025fedsn}. In multi-agent coordination, delayed messages can cause inconsistent local beliefs, outdated task allocations, and conflicting actions, especially when satellites coordinate under rapidly changing mission contexts or time-critical events \cite{handley2018delay,hu2023etcn}. In collaborative inference and tracking, latency reduces the value of exchanged features, tracklets, or state estimates, because the underlying scene or network state may already have evolved by the time those messages arrive \cite{rfc4838,cen2024satflow,yin2025oec}. As a result, practical algorithms must explicitly account for staleness, for example through asynchronous updates, confidence decay, bounded-delay fusion, or value-aware message selection.

\textbf{Bandwidth limitations and message design.}
Bandwidth in on-orbit networks is scarce, time-varying, and mission-dependent. Short contact windows and competing demands from payload downlink, control traffic, and crosslink exchange make it infeasible to share large models, raw imagery, or dense intermediate tensors at high frequency \cite{rfc4838,handley2018delay,yin2025oec}. This has direct implications for algorithm design. Federated learning is pushed toward compressed or structured updates, sparse communication, partial model sharing, or split/federated hybrids \cite{lin2025fedsn,wu2025sflleo}. Multi-agent coordination must often communicate compact intents, bids, or summarized state rather than rich internal representations \cite{choi2009cbba,hu2023etcn}. Collaborative sensing and distributed inference likewise benefit from tracklets, embeddings, early-exit decisions, or uncertainty summaries instead of raw measurements, since the latter quickly exceed realistic contact budgets \cite{branchynet2017,xu2023coinleo,spaceexit2025}. In this sense, message \emph{semantics} matter as much as message size: the most useful shared artifact is often not the most detailed one, but the one that delivers the highest mission value per transmitted bit.

\textbf{Intermittent connectivity and partial participation.}
Unlike terrestrial distributed systems, constellations cannot assume persistent end-to-end connectivity. The DTN literature explicitly models such environments as occasionally connected, frequently partitioned, and dependent on scheduled or opportunistic contacts \cite{rfc4838,rfc9171,rfc9172}. For collaborative algorithms, this means that participation is inherently partial and topology-dependent. In FL, only a subset of satellites may contribute updates during a given synchronization opportunity, making partial participation and contact-aware scheduling unavoidable rather than exceptional \cite{razmi2022groundassisted,lin2025fedsn}. In multi-agent settings, communication graphs may fragment into local neighborhoods, so coordination policies must remain effective under missing messages and changing adjacency relations \cite{handley2018delay,cen2024satflow}. In collaborative inference, intermittent connectivity limits the number of consensus rounds, fusion opportunities, or offloading choices available within a useful time horizon \cite{yin2025oec,xu2023coinleo}. Consequently, algorithms that rely on frequent global synchronization or stable all-to-all communication are often mismatched to realistic on-orbit conditions.

\textbf{Implications for practical algorithm design.}
These constraints suggest several design principles that recur across the three paradigms studied in this survey. First, \emph{asynchrony} is usually more realistic than strict synchrony, because waiting for all nodes to participate often wastes scarce contacts and increases staleness \cite{xie2019async,nguyen2022fedbuff}. Second, \emph{locality-aware collaboration} is often preferable to constellation-wide coordination, since neighborhood- or cluster-level exchange better matches dynamic ISL topologies and bounded message budgets \cite{cen2024satflow,yin2025oec}. Third, \emph{event-driven communication} is often more effective than periodic communication, because it transmits only when the expected utility of sharing exceeds the communication cost \cite{hu2023etcn,branchynet2017}. Fourth, \emph{graceful degradation} is essential: algorithms should continue to operate safely under missing, delayed, or low-confidence communication rather than assuming ideal connectivity.

Overall, latency, bandwidth limits, and intermittent connectivity are not peripheral engineering concerns; they shape the feasible design space of distributed on-orbit AI. Any collaborative method that ignores these factors risks overestimating achievable performance and underestimating operational complexity. For this reason, the following sections discuss federated learning, multi-agent coordination, and collaborative inference not only as algorithmic paradigms, but also as communication-constrained processes whose practical success depends on contact-aware, bandwidth-aware, and delay-tolerant design.

\section{Federated Learning for On-Orbit Intelligence}
Federated learning (FL) enables multiple satellites to collaboratively train or adapt models while keeping raw data onboard \cite{mcmahan2017fedavg,kairouz2021fl}. In satellite constellations, FL is particularly attractive when downlink capacity is limited, when raw sensing data are sensitive, or when onboard models must adapt to changing conditions (illumination, seasons, sensor aging, evolving targets) without relying on continuous ground connectivity \cite{matthiesen2024satfl,razmi2022groundassisted,lin2025fedsn}. Compared with terrestrial FL, on-orbit FL is shaped by intermittent contacts and time-varying topology, heterogeneous compute/energy budgets, and stricter operational requirements for safety, validation, and update governance \cite{razmi2022groundassisted,thangavel2024taso}.

\begin{table}[H]
\caption{Systematic overview of FL for on-orbit intelligence in satellite constellations. The table maps representative constellation-oriented FL works to the taxonomy dimensions (architecture/timing/trust) and highlights the primary on-orbit constraints addressed.}
\label{tab:fl_systematic}
\centering
\footnotesize
\resizebox{\textwidth}{!}{%
\begin{tabular}{p{3.4cm} p{1.7cm} p{2.0cm} p{5.4cm} p{4.8cm} p{2.2cm}}
\hline
\textbf{Work} & \textbf{Arch.} & \textbf{Timing} & \textbf{Main on-orbit focus} & \textbf{Trust / privacy model} & \textbf{Shared artifact} \\
\hline
Ground-assisted FL \cite{razmi2022groundassisted} 
& Centralized (GS) 
& Async (staleness-aware) 
& Designs an asynchronous procedure based on FedAvg to cope with sporadic satellite--GS visibility and heterogeneous system conditions; improves convergence under partial participation. 
& Benign faults / intermittency (no explicit adversary model). 
& Model updates / deltas \\
\hline
Energy-aware scheduling for satellite FL \cite{razmi2024energy} 
& Centralized/Hybrid 
& Orbit-/schedule-aware 
& Schedules local computation time to minimize battery usage (e.g., eclipse periods) while preserving convergence speed; targets long mission lifetime and SWaP-C constraints. 
& Benign (resource reliability and aging). 
& Model updates (standard FL payload) \\
\hline
FedSN \cite{lin2025fedsn} 
& Centralized (GS) 
& Pseudo-sync (staleness compensation) 
& Addresses three LEO-specific bottlenecks: heterogeneous compute/memory, limited uplink rate, and model staleness; uses sub-structure scheme for heterogeneous local training and pseudo-synchronous aggregation. 
& Benign (system heterogeneity / staleness). 
& Structured/partial model updates \\
\hline
FedLEO \cite{zhai2024fedleo} 
& Decentralized (ISL/P2P) 
& Async / contact-driven 
& Offloading-assisted decentralized FL to mitigate central-server bottlenecks; targets straggler effects and statistical heterogeneity under LEO topology and intermittent contacts. 
& Benign (no explicit adversary). 
& Neighbor model mixing + local updates \\
\hline
HiSatFL \cite{hisatfl2025} 
& Hierarchical 
& Contact-/cluster-aware 
& Hierarchical FL tailored to dynamic satellite networks; emphasizes cross-domain adaptation and privacy-preserving collaboration under resource constraints. 
& Privacy-preserving (honest-but-curious assumptions are typical). 
& Privacy-protected model updates \\
\hline
RAFL \cite{xu2025rafl} 
& Hierarchical 
& Contact-aware (reduces GS interaction) 
& Uses aerial networking + hierarchical aggregation to reduce satellite--GS communication; introduces self-adaptive reputation evaluation for leader selection and robustness; improves convergence and resists poisoning. 
& Byzantine/compromised nodes considered (poisoning defense via reputation). 
& Aggregated regional updates + global update \\
\hline
FedSecure (decentralized key generation + on-orbit aggregation) \cite{elmahallawy2023fedsecure} 
& Hybrid (on-orbit forwarding/aggregation + GS) 
& Orbit-periodic / async bridging 
& Reduces convergence delay by partial global aggregation per orbit; protects against insecure channels via decentralized key generation and functional-encryption-based privacy. 
& Honest-but-curious + eavesdroppers (privacy/integrity threats). 
& Encrypted/secure model updates \\
\hline
SFL-LEO (split + federated) \cite{wu2025sflleo} 
& Hybrid (Sat--GS split + FL aggregation) 
& Async local training during disconnection 
& Combines split learning with FL to accommodate constrained onboard compute and high dynamics; enables local training while disconnected; aggregates client-side sub-models at GS. 
& Benign (resource constraints / intermittency). 
& Client-side sub-model parameters \\
\hline
\end{tabular}%
}
\end{table}

\subsection{Topologies and Protocols (Ground-Aggregated, Hierarchical, Decentralized)}
\textbf{Basic formulation.}
Let satellite $i$ maintain a local dataset/stream $\mathcal{D}_i$ and a local objective $F_i(\theta)$ over model parameters $\theta$. A standard global objective is
\begin{equation}
\min_{\theta}\; F(\theta) = \sum_{i=1}^{N} w_i F_i(\theta), 
\quad \text{where } w_i \ge 0,\; \sum_{i=1}^{N} w_i = 1,
\label{eq:fl_global_obj}
\end{equation}
with $w_i$ typically proportional to local data volume and optionally reflecting sensing quality or mission priority \cite{mcmahan2017fedavg,kairouz2021fl}. Satellites perform local training and exchange model-related information (parameters, gradients, deltas, or distilled predictions), while the central design choice is {where} aggregation/coordination happens and {how} updates propagate under a dynamic contact graph.

\textbf{Ground-aggregated FL.}
A ground station (or ground cloud) acts as the aggregator: it distributes a reference model, satellites train locally, and upload updates during downlink windows; the ground then aggregates received updates and releases the next model version. This topology naturally supports centralized governance and validation, and is widely adopted in satellite FL studies to accommodate limited onboard resources \cite{razmi2022groundassisted,matthiesen2024satfl}. Its limitations are long update latency, restricted contact windows, downlink bottlenecks, and persistent partial participation (only satellites with timely contacts contribute to a given cycle), which can amplify staleness and bias \cite{razmi2022groundassisted,lin2025fedsn}.

\textbf{Hierarchical FL.}
Hierarchical FL introduces intermediate aggregators (e.g., cluster-head satellites or region leaders). Satellites first aggregate within local groups via ISLs, then forward summarized updates upward (to a higher-tier satellite or the ground), reducing downlink load and exploiting more frequent ISL opportunities \cite{hisatfl2025,xu2025rafl}. Hierarchical designs are particularly relevant when satellite--ground visibility is sparse and when leader selection and trust must be managed explicitly \cite{xu2025rafl}. Key trade-offs include additional compute/energy burden at leaders, sensitivity to leader failures, and the need for robust/secure intra-cluster aggregation.

\textbf{Decentralized FL.}
Decentralized FL removes the assumption of an always-available central aggregator. Each satellite maintains a local model $\theta_i$ and exchanges information with neighbors when contacts occur, often through consensus/gossip mixing combined with local SGD \cite{lian2017dpsgd,zhai2024fedleo}. One common abstraction is a ``mix-then-train'' update:
\begin{equation}
\tilde{\theta}_i^{(t)} = \sum_{j \in \mathcal{N}_i(t)\cup\{i\}} a_{ij}(t)\,\theta_j^{(t)}, 
\qquad 
\theta_i^{(t+1)} = \tilde{\theta}_i^{(t)} - \eta\, g_i\!\big(\tilde{\theta}_i^{(t)}\big),
\label{eq:decentralized_mix_train}
\end{equation}
where $\mathcal{N}_i(t)$ is the set of neighbors reachable at time $t$, $a_{ij}(t)$ are mixing weights (row-stochastic over the available neighborhood), and $g_i(\cdot)$ denotes a stochastic gradient computed from $\mathcal{D}_i$ \cite{lian2017dpsgd}. Decentralized FL can improve resilience when ground contact is sparse, and has been explored for LEO settings with computation/communication offloading and dynamic topology \cite{zhai2024fedleo}. However, it is more sensitive to topology changes, staleness, and trust assumptions, and typically requires additional mechanisms for stability and quality control.

\textbf{What to transmit.}
On-orbit FL rarely transmits raw data; the communication budget often motivates exchanging compact model information: parameter differences, low-rank/structured updates, quantized or sparsified updates, and (in some cases) distilled predictions such as logits on shared proxy inputs \cite{konecny2016comm,reizizadeh2020fedpaq,alistarh2017qsgd,lin2018dgc,sattler2020stc,jeong2018feddistill,li2019fedmd}. These choices trade accuracy and convergence against bandwidth/energy, and interact with intermittency because short contact windows favor small, quickly verifiable payloads.

\textbf{Hybrid and split protocols.}
Many constellation designs are hybrid: satellites may aggregate locally via ISLs most of the time, while the ground periodically synchronizes, validates, and authorizes releases \cite{razmi2022groundassisted,xu2025rafl}. Another increasingly relevant pattern is to combine FL with model splitting or on-orbit forwarding/partial aggregation to cope with constrained satellite compute and sporadic satellite--ground connectivity \cite{wu2025sflleo,elmahallawy2023fedsecure}.

\subsection{Intermittency and Efficiency (Asynchrony, Partial Participation, Compression/Scheduling)}
\textbf{Asynchrony and stale updates.}
Strictly synchronous rounds are difficult in constellations because contact opportunities are intermittent and topology changes with orbital motion. Asynchronous FL allows satellites to send updates whenever contacts occur, while the coordinator incorporates updates upon arrival, typically with staleness-aware weighting or bounded-delay rules \cite{xie2019async,razmi2022groundassisted}. Buffered asynchronous aggregation (e.g., aggregating a buffer of $K$ updates before a server step) can reduce straggler sensitivity while remaining compatible with privacy primitives such as secure aggregation \cite{nguyen2022fedbuff}. In satellite settings, pseudo-synchronous or staleness-compensated aggregation has been proposed to better match periodic visibility patterns and mitigate stale contributions \cite{lin2025fedsn}.

\textbf{Partial participation and participant selection.}
Only a subset of satellites can participate in any interval due to contact availability, duty cycling, and resource constraints. Participant selection can be treated as a constrained optimization problem balancing utility, timeliness, and resource cost. Terrestrial FL has developed resource-aware selection strategies (e.g., FedCS) and utility-aware selection (e.g., Oort), which are conceptually aligned with contact- and energy-limited satellite participation \cite{nishio2018fedcs,lai2021oort}. In constellations, selection policies should additionally incorporate predicted contact windows, remaining energy, compute load, and mission priority, and should avoid systematic bias across orbital planes/shells over long time horizons \cite{razmi2024energy,lin2025fedsn}.

\textbf{Communication reduction.}
Communication is often the dominant bottleneck. Classic techniques include structured/sketched updates, quantization, sparsification (e.g., top-$k$), and error feedback \cite{konecny2016comm,alistarh2017qsgd,lin2018dgc}. FL-specific schemes such as periodic averaging with quantization (FedPAQ) formalize the computation--communication trade-off \cite{reizizadeh2020fedpaq}. In highly non-IID regimes, compression must be designed carefully because aggressive sparsification/quantization can amplify client drift; FL-tailored compression such as STC explicitly targets non-IID + partial participation regimes \cite{sattler2020stc}.

\textbf{Scheduling with contact plans.}
Orbital dynamics and network schedules provide predictable contact windows. FL can exploit this structure by aligning aggregation opportunities with expected satellite--ground/ISL contacts, allocating more local computation when the next contact is far, and reducing compute load near imminent synchronization \cite{razmi2022groundassisted,wu2025sflleo}. Energy-aware schedulers further highlight that ``when to train'' can be as important as ``how to aggregate'' for long mission lifetimes \cite{razmi2024energy}.

\textbf{Energy- and compute-aware training.}
Training competes with payload operations, attitude control, routing, and housekeeping. Practical designs adapt local training frequency, batch size, and precision to available power and compute, and may restrict training to designated duty-cycle windows \cite{razmi2024energy,lin2025fedsn}. Model structuring (e.g., updating only small adapters/heads) can also reduce both compute and communication while lowering operational risk.

\textbf{Local sanity checks before upload.}
Before transmitting updates, satellites can apply lightweight checks to reduce the probability of propagating corrupted or unstable updates (e.g., norm clipping, divergence detection, simple statistical consistency tests). These checks complement robust aggregation methods designed to tolerate abnormal or adversarial updates \cite{blanchard2017krum,yin2018byzantine}.

\subsection{Robustness, Security, and Safe Updates (Non-IID, Poisoning, Rollback/Governance)}
\textbf{Non-IID data and heterogeneity.}
Cross-satellite data heterogeneity can slow convergence and bias global models. Beyond FedAvg \cite{mcmahan2017fedavg}, widely used heterogeneity-mitigation approaches include proximal regularization (FedProx), variance-reduction against client drift (SCAFFOLD), and normalization to correct objective inconsistency (FedNova) \cite{li2020fedprox,karimireddy2020scaffold,wang2020fednova}. Personalization is often more appropriate than a single global model in constellations with persistent domain differences; representative strategies include global-base + local-head (FedPer), bilevel personalization (pFedMe), and personalization for fairness/robustness (Ditto) \cite{arivazhagan2019fedper,dinh2020pfedme,li2021ditto}. Constellation-focused surveys and systems similarly emphasize that non-IID and system heterogeneity are first-order design constraints for satellite FL \cite{matthiesen2024satfl,lin2025fedsn}.

\textbf{Concept drift and continual adaptation.}
On-orbit data streams exhibit drift due to seasonality, illumination changes, sensor aging, and evolving targets, often with sparse or delayed labels. This motivates drift-aware evaluation, periodic re-synchronization, and controlled continual learning, but with stronger validation gates for safety-critical tasks \cite{tran2026conceptdrift,thangavel2024taso}.

\textbf{Robustness to faults and unreliable participants.}
Radiation-induced transient faults and compute/memory errors can surface as abnormal or inconsistent updates. Robust aggregation rules (e.g., Krum / trimmed-mean/median families) and consistency filtering help even under benign faults, and become essential under stronger threat models \cite{blanchard2017krum,yin2018byzantine,souvatzoglou2024seu}. Satellite-focused designs also incorporate leader selection/reputation mechanisms to mitigate unreliable or compromised participants in hierarchical settings \cite{xu2025rafl}.

\textbf{Threats and poisoning.}
If a satellite is compromised, it may attempt model poisoning or backdoor insertion. Model-replacement style backdoors demonstrate that a single malicious participant can implant hidden behaviors while maintaining apparent task accuracy, underscoring the need for explicit threat modeling and robust defenses \cite{bagdasaryan2020backdoor,blanchard2017krum,yin2018byzantine}.

\textbf{Privacy and confidentiality.}
Although raw data are not shared, model updates can leak information. Secure aggregation hides individual updates from the aggregator, while differential privacy bounds information leakage at the cost of accuracy and additional compute/communication overhead \cite{bonawitz2017secureagg,mcmahan2018dprnn}. In satellite networks, privacy/security concerns also include eavesdropping over insecure links and curious relays, motivating cryptography-assisted designs and on-orbit aggregation variants \cite{elmahallawy2023fedsecure}. In practice, secure on-orbit FL requires confidentiality and integrity to be handled jointly rather than separately. Protecting update contents through secure aggregation or cryptographic mechanisms is insufficient unless paired with integrity checks, robust aggregation, authenticated model versions, and rollback-capable release governance.

\textbf{Safe deployment, versioning, and rollback.}
In-orbit model updates should be treated as controlled, signed, and versioned artifacts. A practical approach is staged rollout (pilot satellites first), telemetry-based monitoring, and automatic rollback to a known-safe baseline upon detected regressions \cite{thangavel2024taso}. These mechanisms are critical for maintaining mission safety envelopes when models adapt under non-IID drift and intermittent connectivity.

\textbf{Operational integration.}
Finally, on-orbit FL must fit within mission operations: coordinating training with payload schedules and network policies, logging for auditing/debugging, and explicitly defining when adaptation is permitted (and when it is prohibited). In practice, these operational constraints often determine feasibility and the acceptable aggressiveness of in-orbit adaptation \cite{thangavel2024taso,razmi2022groundassisted}.

\begin{table}[H]
\caption{Technique-level index for on-orbit federated learning: challenges, representative methods, and satellite-oriented adaptations.}
\label{tab:fl_tech_index}
\centering
\footnotesize
\resizebox{\textwidth}{!}{%
\begin{tabular}{p{3.4cm} p{5.8cm} p{6.7cm} p{3.0cm}}
\hline
\textbf{Challenge} & \textbf{Representative techniques} & \textbf{Satellite-oriented instantiation} & \textbf{Key refs.} \\
\hline
Intermittency / staleness 
& Async FL; buffered async; staleness-aware weighting 
& Contact-driven uploads; pseudo-synchrony aligned with visibility; bounded-delay update admission
& \cite{xie2019async,nguyen2022fedbuff,razmi2022groundassisted,lin2025fedsn} \\

Communication bottleneck
& Quantization; sparsification; periodic averaging + quantization; non-IID-aware compression
& Transmit deltas/partial modules; quick-to-verify payloads for short windows
& \cite{konecny2016comm,alistarh2017qsgd,lin2018dgc,reizizadeh2020fedpaq,sattler2020stc} \\

System heterogeneity (compute/memory)
& Partial/structured training; split/federated hybrids
& Train client-side sub-models; aggregate only trainable subsets; match onboard capability
& \cite{lin2025fedsn,wu2025sflleo} \\

Statistical heterogeneity (non-IID)
& Proximal regularization; control variates; normalized aggregation
& Stabilize local drift when long gaps exist; improve fairness across orbital planes/domains
& \cite{li2020fedprox,karimireddy2020scaffold,wang2020fednova} \\

Personalization
& Global-base + local layers; bilevel personalization; robust personalization
& Per-orbit/plane local specialization with safe baselines and limited adaptation scope
& \cite{arivazhagan2019fedper,dinh2020pfedme,li2021ditto} \\

Robustness to faulty/Byzantine updates
& Byzantine-robust aggregation; outlier filtering
& Fault-induced abnormal updates treated similarly to malicious outliers; hierarchical reputation
& \cite{blanchard2017krum,yin2018byzantine,xu2025rafl} \\

Privacy/confidentiality
& Secure aggregation; differential privacy; crypto-assisted designs
& Secure on-orbit aggregation/forwarding under insecure channels; privacy-budget-aware scheduling
& \cite{bonawitz2017secureagg,mcmahan2018dprnn,elmahallawy2023fedsecure} \\

Poisoning/backdoors
& Threat modeling; robust aggregation; governance
& Staged rollout + telemetry + rollback; reputation/attestation where feasible
& \cite{bagdasaryan2020backdoor,blanchard2017krum,thangavel2024taso} \\
\hline
\end{tabular}%
}
\end{table}

\section{Multi-Agent Coordination for Constellation Autonomy}

Multi-agent coordination treats each satellite as an autonomous decision-maker that acts under local constraints while contributing to constellation-level objectives. Relative to single-spacecraft autonomy, constellations introduce strong coupling through shared resources (coverage, downlink, spectrum, compute, and power), time-varying communication graphs, and system-level safety requirements (e.g., collision avoidance and coordinated maneuvering) \cite{oliehoek2016decpomdp,thangavel2024taso,wang2026agent}. These properties naturally induce decentralized decision-making under partial observability and intermittent communication, where both the {content} and {timing} of information exchange become part of the control policy \cite{handley2018delay,rfc4838}. Accordingly, coordination methods must address non-stationarity, heterogeneous capabilities, delayed and missing messages, and the need for robust long-term operation over multi-year mission lifetimes \cite{hernandezleal2019marl_survey,gronauer2022marl_survey}. The transition from single-satellite AI to constellation-scale multi-agent autonomy is not merely a matter of scaling up node count. It introduces distributed partial observability, coupled decision spaces, dynamic communication graphs, and safety-critical interdependence among satellites, all of which fundamentally reshape the algorithm design problem.

\subsection{Core Tasks (Coverage, Scheduling, Formation, Collision Avoidance)}
\textbf{Coverage and persistent monitoring.}
A central objective is to maximize spatiotemporal coverage and revisit performance for regions of interest under viewing geometry, duty-cycle limits, and energy constraints. In Earth observation and related missions, coverage objectives are typically realized through time-windowed opportunities and priority changes (e.g., disaster response), making coordination necessary to reduce redundant sensing and avoid missed opportunities across orbital planes \cite{lemaitre2002agile,ferrari2025ssp}.

\textbf{Tasking and observation scheduling.}
Constellation scheduling assigns observation tasks to satellites and selects sensing modes, pointing commands, and time allocations subject to slew limits, storage, and downlink constraints. This remains a core operations problem with many variants (imaging and communication tasks; uncertainty; multiple objectives), and is tightly coupled to onboard processing and product delivery latency \cite{lemaitre2002agile,ferrari2025ssp}. Decentralized or auction-based scheduling is especially relevant when operators seek scalable, interpretable coordination under intermittent connectivity \cite{picard2022eoscsp,choi2009cbba}.

\textbf{Resource allocation and network-aware coordination.}
Autonomous constellations must allocate shared resources such as ISL time, downlink windows, onboard compute slots, and power. Network-aware coordination further includes routing and store-and-forward decisions in DTN-like settings, and value-aware prioritization of which products to transmit when contacts appear \cite{rfc4838,araniti2015cgr,handley2018delay}. These decisions become more complex under heterogeneous payloads and roles, and are increasingly studied in the context of large LEO networks and scalable contact-aware planning \cite{cen2024satflow}.

\textbf{Formation flying and reconfiguration.}
For missions involving distributed apertures or coordinated sensing geometry, satellites must maintain relative configurations and may need to reconfigure or retask under fuel and safety constraints. Formation flying introduces coupled dynamics, relative navigation uncertainty, and communication delays, motivating distributed control and consensus-style coordination \cite{scharf2003ffg,scharf2004ffc,dimauro2018gnc_survey,olfati2007consensus}.

\textbf{Collision avoidance and risk management.}
Collision avoidance (COLA) is safety-critical and increasingly central as the orbital environment becomes more congested. At constellation scale, maneuvers affect coverage, scheduling, and network topology, and coordination can be required to avoid conflicting maneuvers and to manage risk under uncertain conjunction data and delayed updates \cite{nasa2023ca2,newman2022amos,sorge2025cola}. Recent studies also highlight the need for operationally viable decision-planning and coordination protocols for autonomous mitigation \cite{bourriez2023collision,autoca_esa}.

\subsection{Algorithms (MARL, Distributed Planning/Allocation)}
\textbf{Formal decision models and CTDE.}
Many coordination problems can be formalized as decentralized decision-making under partial observability (e.g., Dec-POMDP-style settings), where each satellite observes locally and must act with limited communication \cite{oliehoek2016decpomdp}. In practice, multi-agent reinforcement learning (MARL) is frequently adopted when environment models are incomplete and objectives are long-horizon. A common paradigm is centralized training with decentralized execution (CTDE), enabling the use of global information and simulation during training while restricting policies to local observations (and limited messages) at runtime \cite{hernandezleal2019marl_survey,gronauer2022marl_survey}. Representative CTDE families include actor--critic methods (e.g., MADDPG; COMA) and value-decomposition/value-factorization (e.g., VDN; QMIX) \cite{lowe2017maddpg,foerster2018coma,sunehag2017vdn,rashid2018qmix}. For constellation settings with variable neighborhood structure, graph-based policy representations are a natural choice to generalize across changing connectivity \cite{jiang2018gcrl}.

\textbf{Communication-aware coordination.}
Because communication is constrained and intermittent, coordination policies must account for delays, packet loss, and topology changes. A substantial MARL literature treats communication as a learnable component—deciding {what} to communicate, {to whom}, and {when}. Representative mechanisms include continuous message passing (CommNet), differentiable inter-agent learning (DIAL), gating/when-to-communicate models (IC3Net), targeted communication (TarMAC), and graph-attention scheduling of communication edges (MAGIC) \cite{sukhbaatar2016commnet,foerster2016dial,singh2019ic3net,das2019tarmac,niu2021magic}. Event-triggered communication policies provide an additional lens that aligns well with contact- and bandwidth-limited satellite settings \cite{hu2023etcn}. When links are unavailable, robust coordination benefits from explicit uncertainty tracking, conservative rules, and policies that degrade gracefully under missing information.

\textbf{Distributed planning and task allocation.}
Many constellation problems can be expressed as distributed planning with explicit constraints and interpretable objectives. Market-based and auction-based allocation, including consensus-based bundle assignment, can be implemented with limited message exchange (bids/utilities) while naturally handling heterogeneous capabilities \cite{choi2009cbba,picard2022eoscsp}. Distributed constraint optimization (DCOP) provides a general coordination abstraction with both complete and approximate algorithms, and has been widely surveyed \cite{fioretto2018dcop_survey,modi2005adopt}. These methods are attractive for operations because constraints are explicit, decision logic is auditable, and integration with mission rules (e.g., priority policies) is straightforward.

\textbf{Distributed control and optimization.}
For formation flying, networked control, and some allocation problems, distributed control and optimization can offer predictable behavior with explicit constraints. Consensus and cooperative control foundations are well established for dynamic graphs and delays \cite{olfati2007consensus}. Distributed and hierarchical model predictive control (MPC/DMPC) provides a powerful framework for explicitly enforcing state/input constraints and coordinating coupled subsystems, with several foundational surveys and roadmaps \cite{scattolini2009mpc,christofides2013dmpc,negenborn2014dmpc}. In constellation contexts, these approaches are often combined with contact-aware scheduling and simplified models to remain feasible under onboard resource limits.

\textbf{Hybrid approaches.}
Operational systems are often hybrid: learning-based modules provide fast heuristics, prioritization, or policy initialization, while optimization-based layers enforce hard constraints and safety envelopes. Hybrid designs can also separate timescales—slow planning/allocation via optimization and fast reaction via learned policies—while retaining runtime monitors and conservative fallbacks \cite{thangavel2024taso}.

\subsection{Safety and Operations (Constraints, Runtime Assurance, Human-on-the-Loop)}
\textbf{Hard constraints and safe decision-making.}
Constellation autonomy must respect hard constraints (collision avoidance, attitude and slew limits, thermal/power budgets, link schedules, fuel usage). Rather than treating constraints as soft penalties, many approaches incorporate constrained RL formulations and safety layers/filters that modify proposed actions to remain within admissible sets \cite{garcia2015safeRL,achiam2017cpo,alshiekh2018shield,wabersich2021safetyfilter}. These ideas are directly relevant to maneuver planning and risk-limited autonomy, where unsafe actions may be mission-ending.

\textbf{Runtime assurance and monitoring.}
Even when policies are learned or adaptive, on-orbit operation typically requires runtime assurance (RTA): monitors that check safety-relevant properties online and switch to verified fallback behaviors when risk thresholds are exceeded \cite{cofer2020rta,hobbs2023rta}. In satellite operations, this mindset aligns with conservative fault protection and staged autonomy, and complements conjunction assessment best practices and operator coordination protocols \cite{nasa2023ca2,thangavel2024taso}.

\textbf{Validation, staging, and operational governance.}
Constellation autonomy is commonly introduced gradually: simulation validation (including digital twins), hardware-in-the-loop tests when available, and staged rollout from advisory modes to supervised autonomy and finally to higher autonomy under well-defined envelopes \cite{thangavel2024taso}. Governance also includes versioning, audit trails, and rollback policies to maintain predictable behavior under drift and evolving conditions.

\textbf{Human-on-the-loop coordination.}
Many missions require human operators to inspect, approve, or veto safety-relevant actions (especially maneuvers). Human-on-the-loop designs emphasize transparency of intent and uncertainty, appropriate levels of automation, and interfaces that summarize fleet-level decisions because operators cannot track each satellite in detail \cite{sheridan1992telerobotics,parasuraman2000automation,endsley1995sa}. In constellation settings, this often motivates explainable summaries, conservative defaults, and mechanisms to reconcile local autonomy with centralized operational policies.

\textbf{Long-term robustness.}
Constellations operate for years while payloads degrade, orbits evolve, and the operational environment changes. Multi-agent autonomy therefore benefits from conservative adaptation, strong telemetry for auditing, and clear separation between components that may adapt and those that must remain stable and certifiable \cite{thangavel2024taso}.

\begin{table}[H]
\caption{Systematic overview of multi-agent coordination methods for constellation autonomy.}
\label{tab:ma_systematic}
\centering
\footnotesize
\resizebox{\textwidth}{!}{%
\begin{tabular}{p{3.3cm} p{4.9cm} p{4.8cm} p{3.3cm} p{4.7cm} p{3.2cm}}
\hline
\textbf{Method family} & \textbf{Representative algorithms} & \textbf{Typical constellation tasks} & \textbf{What is shared} & \textbf{Fit to orbital constraints} & \textbf{Key refs.} \\
\hline
CTDE MARL (value/actor-critic)
& MADDPG; COMA; VDN/QMIX
& Scheduling/coverage; routing policies; reactive tasking
& Learned messages / local obs summaries
& Strong in simulation; execution decentralized; needs robustness to delay and non-stationarity
& \cite{lowe2017maddpg,foerster2018coma,sunehag2017vdn,rashid2018qmix} \\

Learned communication in MARL
& CommNet; DIAL; IC3Net; TarMAC; MAGIC
& Distributed coordination under partial observability
& Differentiable messages; gated communication
& Enables ``what/when/to whom'' policies; aligns with bandwidth limits (event-trigger)
& \cite{sukhbaatar2016commnet,foerster2016dial,singh2019ic3net,das2019tarmac,niu2021magic,hu2023etcn} \\

Auction / market-based allocation
& CBBA; distributed auction scheduling
& Observation tasking; downlink allocation
& Bids/utilities; local plans
& Interpretable; works with sparse comm; good for heterogeneity and auditable ops
& \cite{choi2009cbba,picard2022eoscsp} \\

DCOP (distributed constraint optimization)
& ADOPT; DCOP frameworks
& Constraint-heavy scheduling/resource allocation
& Constraint messages / cost tables
& Explicit constraints; can trade optimality for comm; supports auditability
& \cite{modi2005adopt,fioretto2018dcop_survey} \\

Distributed control / consensus / DMPC
& Consensus control; (D)MPC surveys
& Formation flying; reconfiguration
& State/trajectory/constraints
& Predictable + constraint handling; sensitive to delays; may need simplified models
& \cite{olfati2007consensus,scattolini2009mpc,christofides2013dmpc,negenborn2014dmpc} \\

Safety layers \& runtime assurance
& Safe RL; CPO; shielding; predictive safety filters; RTA
& Maneuver planning; risk-limited autonomy
& Safety monitors / filters
& Enforces hard constraints; supports staged autonomy and fallbacks
& \cite{garcia2015safeRL,achiam2017cpo,alshiekh2018shield,wabersich2021safetyfilter,cofer2020rta,hobbs2023rta} \\

COLA / conjunction operations (governance)
& Best practices; autonomous COLA concepts
& Collision avoidance and risk management
& Risk thresholds; maneuver intents
& Mission-critical; human-on-the-loop common; requires conservative governance
& \cite{nasa2023ca2,newman2022amos,sorge2025cola,autoca_esa,bourriez2023collision} \\
\hline
\end{tabular}%
}
\end{table}

\section{Collaborative Sensing and Distributed Inference}

Collaborative sensing and distributed inference focus on constellation-level perception and estimation by sharing information across satellites and distributing parts of the inference pipeline. This differs from federated learning (training/adaptation) and from multi-agent coordination (action selection): here the emphasis is on {inference-time collaboration}---fusing observations, exchanging intermediate results, and exploiting the constellation as a distributed computing fabric to obtain faster or more reliable situational understanding. In orbit, this collaboration is constrained by intermittent ISLs, time-varying topology, heterogeneous sensors and compute, and strict latency requirements for time-critical tasks (e.g., conjunction risk, event detection) \cite{rfc4838,handley2018delay,yin2025oec}.

\subsection{Multi-Satellite Fusion and Tracking (Consensus/Information Filters)}
Multi-satellite fusion becomes essential when a single observation is insufficient, ambiguous, or too noisy, and when combining multiple views improves uncertainty reduction and robustness. Representative use cases include cooperative target tracking and orbit determination, conjunction assessment and SSA, and multi-view Earth observation products \cite{barshalom2001estimation,khaleghi2013fusionreview}.

\textbf{Distributed filtering and track-to-track fusion.}
A common approach maintains a local posterior/track on each satellite and exchanges compact sufficient statistics (rather than raw measurements) \cite{barshalom2001estimation}. For Gaussian track fusion, it is convenient to use the information form, where each node $i$ represents its estimate with an information matrix/vector $(\mathbf{Y}_i,\mathbf{y}_i)=(\mathbf{P}_i^{-1},\mathbf{P}_i^{-1}\hat{\mathbf{x}}_i)$. Under conditional independence assumptions, information pairs add; however, in peer-to-peer settings cross-correlations are often unknown, and naive fusion can lead to overconfident estimates (``data incest'') \cite{battistelli2014klavg}. A classical conservative remedy is covariance intersection (CI), which fuses two information pairs via a convex combination
\begin{equation}
(\mathbf{Y},\mathbf{y})=\omega(\mathbf{Y}_1,\mathbf{y}_1) + (1-\omega)(\mathbf{Y}_2,\mathbf{y}_2),\quad \omega\in[0,1],
\end{equation}
guaranteeing consistency without requiring the cross-correlation \cite{julier1997ci,hurley2002ci}. This CI/KL-average viewpoint connects naturally to distributed Bayesian fusion based on information-theoretic averaging (e.g., Kullback--Leibler averaging and consensus on posteriors) \cite{battistelli2014klavg}.

\textbf{Consensus-based distributed estimators.}
Consensus filters provide a generic mechanism for distributed averaging on time-varying graphs and are widely used to build distributed Kalman/information filters. Early influential work decomposed centralized filtering into local micro-filters driven by consensus over measurements and inverse covariances \cite{olfatisaber2005dkf,olfatisaber2005consensusfilters}. Subsequent developments include information-consensus filters that combine information filtering with consensus iterations and explicitly analyze track-to-track statistics \cite{casbeer2009icf}. More recent work also connects distributed Kalman filtering to consensus optimization viewpoints with stability/performance guarantees \cite{ryu2023dkfopt}. In constellation settings, these methods must additionally confront intermittent contacts and limited consensus iterations per window, motivating bounded-round consensus, timeouts, and staleness-aware weighting \cite{rfc4838,handley2018delay}.

\textbf{Multi-object tracking and data association.}
For multi-target tracking, fusion must handle data association, false alarms, and track management. When bandwidth is limited, satellites typically exchange compact tracklets (state, covariance, quality flags) or compressed descriptors rather than full sensor data \cite{barshalom2001estimation}. Random finite set (RFS) formulations provide a principled Bayesian framework for multi-object tracking and have also been adapted to consensus/distributed settings, including consensus labeled RFS filters \cite{mahler2007fisst,fantacci2015rfs,wei2018debris}. These approaches are attractive for SSA-like scenarios where reporting track uncertainty and avoiding double counting are critical.

\begin{figure}
    \centering
    \includegraphics[width=\linewidth]{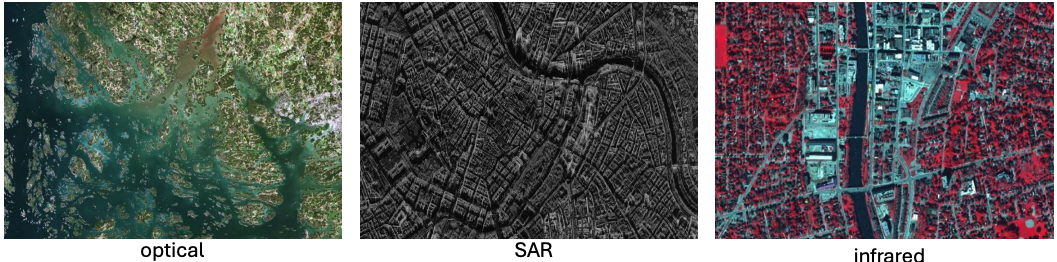}
    \caption{Example Remote Sensing Images from Satellite with Different Sensors. }
    \label{fig:images}
\end{figure}

\textbf{Multi-view and heterogeneous sensing.}
Fusion is not limited to identical sensors: constellations may combine optical, SAR, infrared, GNSS reflectometry, and RF sensing. Examples images are illustrated in Figure \ref{fig:images}.  Heterogeneous fusion requires calibration and temporal alignment and benefits from explicit uncertainty modeling so that contributions can be weighted by geometry and sensor quality \cite{khaleghi2013fusionreview}. In practice, heterogeneity also influences what is shared (raw measurements vs. features vs. tracks), because the latter determines both the achievable latency and the robustness to intermittent links.

\subsection{Collaborative Representation and Inference (Self-Supervised, Split/Early-Exit)}
\textbf{Representation sharing under scarce labels.}
When labels are sparse or delayed, collaborative representation learning can improve generalization by aligning features across satellites without transmitting raw imagery. Self-supervised learning (SSL) is particularly relevant for remote sensing because it exploits abundant unlabeled data and has been comprehensively reviewed and benchmarked in the EO domain \cite{wang2022sslrs,tao2022ssrsfeature}. In constellation settings, satellites may exchange compact prototypes, reference embeddings, or statistics to encourage cross-node consistency under non-IID conditions, while keeping bandwidth low.

\textbf{Collaborative self-supervision under non-IID data.}
Non-IID distributions can cause representations to drift apart across orbital planes, seasons, and sensor conditions. Alignment can be encouraged using contrastive or representation-regularization objectives across participants (e.g., model-contrastive or federated contrastive schemes), which conceptually match ``share representations, not raw data'' designs \cite{li2021moon,dong2021fedmoco}. In orbit, such collaboration is often opportunistic (only during contact windows) and must be robust to partial participation.

\textbf{Split inference and partitioned computation.}
Distributed inference can split a model across satellites or between satellite and ground, executing early layers to generate an intermediate representation that is transmitted for completion elsewhere. A minimal abstraction is
\begin{equation}
\mathbf{z}=f_{1:k}(\mathbf{x}),\qquad \hat{\mathbf{y}}=f_{k+1:L}(\mathbf{z}),
\end{equation}
where $\mathbf{z}$ can be quantized/compressed to match link budgets. Dynamic partitioning between edge and cloud has been studied extensively in terrestrial edge systems (e.g., layer-wise partitioning to minimize end-to-end latency/energy) \cite{kang2017neurosurgeon}. In satellite edge computing, recent works explicitly study multi-satellite collaborative inference, including pipeline-style model splitting and satellite selection under LEO topology \cite{xu2023coinleo}, as well as gain-aware scheduling for multi-exit models in resource-constrained in-orbit settings \cite{zhang2025satcooper}.

\textbf{Early-exit inference and adaptive depth.}
Early-exit models attach intermediate classifiers so that ``easy'' samples exit early, reducing compute and often enabling event-driven communication (only uncertain cases are forwarded for deeper inference or collaborative verification) \cite{branchynet2017,spaceexit2025}. A comprehensive survey of early-exit DNNs further systematizes design and training choices, which are increasingly relevant for latency-critical on-orbit analytics \cite{rahmath2024earlyexit}. In the satellite context, multi-exit DNNs have also been explored for cooperative inference where exit-point selection interacts with offloading and link states \cite{zhang2025satcooper}.

\textbf{Collaborative ensembles and expert selection.}
Another practical pattern is to distribute specialists (experts) across the constellation (e.g., day/night, land/ocean, latitude bands) and query the most suitable expert based on metadata and uncertainty. This can improve robustness under persistent domain shift while avoiding a single monolithic model, but requires careful control of communication frequency and validation of received outputs under intermittent contacts \cite{yin2025oec}.

\subsection{Fault Tolerance and Cross-Layer Co-Design (Offloading, AI-for-Networking)}
\textbf{Fault tolerance under transient failures.}
Space environments introduce transient faults, resets, and silent corruptions. Collaborative inference should degrade gracefully when nodes disappear or messages are corrupted, using bounded trust in received messages, redundancy (multi-satellite confirmation), consistency checks on shared statistics, and conservative fusion when correlations are uncertain \cite{julier1997ci,hurley2002ci,wang2023mars,souvatzoglou2024seu}. For split pipelines, practical safeguards include checkpointing and the ability to revert to local-only inference if collaboration becomes unavailable.

\textbf{Compute offloading and scheduling.}
Collaborative inference is tightly linked to offloading decisions: process locally, forward to a neighbor, or downlink to the ground. Orbital edge computing surveys emphasize that feasibility depends on contact windows, hop delays, and the energy trade-off between compute and transmission \cite{yin2025oec,wu2023oec}. Recent satellite-focused studies formulate and evaluate collaborative inference policies that jointly decide satellite selection, partitioning, and scheduling under LEO network constraints \cite{xu2023coinleo,zhang2025satcooper}.

\textbf{Cross-layer co-design: AI-for-networking and networking-for-AI.}
Networking determines which collaboration patterns are feasible, while AI workloads shape network demand. AI-for-networking uses learning/optimization to improve routing, contact utilization, and planning in large LEO networks \cite{handley2018delay,cen2024satflow}. Networking-for-AI, conversely, prioritizes messages that maximize inference utility (e.g., uncertainty reduction, consensus completion, or timely alerts), rather than treating all packets equally. This co-design view is central when contact windows are short and when system-level objectives (latency, energy, reliability) dominate raw model accuracy \cite{rfc4838,yin2025oec}.

\textbf{Event-driven communication.}
Event-driven policies transmit only when the value of communication is high, such as when confidence exceeds a threshold (triggering alerts), or when uncertainty is high (requesting collaborative verification). Early-exit and multi-exit architectures naturally support such designs by activating deeper or distributed computation only when needed \cite{branchynet2017,spaceexit2025,zhang2025satcooper}. Similar principles appear in event-triggered communication for multi-agent learning/control, aligning well with intermittency-limited ISLs \cite{hu2023etcn}.

\textbf{End-to-end reliability and operational integration.}
Finally, collaborative inference should be evaluated end-to-end, including delays, drops, partial participation, and fault injection, because idealized networking can overestimate real mission benefit. Operational integration requires explicit latency targets, defined fallback behaviors, and auditable logging of shared outputs (tracks, alerts, fused estimates) for post-event analysis and safety review \cite{rfc4838,yin2025oec}.

\begin{table}[H]
\caption{Systematic overview of collaborative sensing and distributed inference in constellations.}
\label{tab:collab_inference_systematic}
\centering
\footnotesize
\resizebox{\textwidth}{!}{%
\begin{tabular}{p{3.4cm} p{4.7cm} p{4.6cm} p{4.2cm} p{4.8cm} p{3.0cm}}
\hline
\textbf{Method family} & \textbf{Core idea} & \textbf{What is shared} & \textbf{Contact/timing sensitivity} & \textbf{Typical use cases} & \textbf{Key refs.} \\
\hline
Conservative fusion (CI)
& Fuse without cross-correlation knowledge; ensure consistency
& Information pairs / covariances
& Low iteration; robust under intermittent contacts
& SSA track fusion; reliable estimation under uncertain correlation
& \cite{julier1997ci,hurley2002ci} \\

Information-theoretic averaging (KL-average)
& Consensus on probability densities with stability guarantees
& Density/summary statistics
& Needs exchanges but can be bounded; stable under partial graphs
& Distributed state estimation / tracking
& \cite{battistelli2014klavg} \\

Consensus Kalman / information filters
& Distributed filtering via consensus iterations
& Measurements/information vectors
& Sensitive to rounds per contact window; requires bounded-round design
& Cooperative tracking; orbit estimation
& \cite{olfatisaber2005dkf,olfatisaber2005consensusfilters,casbeer2009icf,ryu2023dkfopt} \\

RFS-based distributed multi-object tracking
& Bayesian multi-object tracking; consensus variants
& Tracklets / labeled RFS representations
& Exchange compact track summaries; handles association uncertainty
& Debris/RSO tracking; SSA multi-target tracking
& \cite{mahler2007fisst,fantacci2015rfs,wei2018debris} \\

Split inference / partitioned computation
& Compute early layers onboard; transmit embeddings
& Intermediate features / embeddings
& Very contact-aware; choose split point based on latency/bandwidth
& EO event detection; cooperative analytics
& \cite{kang2017neurosurgeon,xu2023coinleo} \\

Early-exit / multi-exit inference
& Exit early for easy samples; trigger collaboration on uncertainty
& Decisions + optional features for hard cases
& Naturally event-driven; reduces comm and compute
& Time-critical detection; bandwidth-limited verification
& \cite{branchynet2017,spaceexit2025,rahmath2024earlyexit,zhang2025satcooper} \\

Cross-layer co-design (networking-for-AI)
& Route/schedule messages by inference utility
& Utility-weighted messages/alerts
& Requires DTN/contact-plan awareness
& Low-latency alerting; uncertainty reduction per bit
& \cite{rfc4838,handley2018delay,yin2025oec,cen2024satflow} \\
\hline
\end{tabular}%
}
\end{table}

\subsection{Representative Datasets and Benchmark Resources}

A practical challenge in distributed on-orbit Space AI is the lack of unified benchmarks that jointly capture constellation networking, onboard compute limits, non-IID sensing, and operational constraints. Nevertheless, several public datasets and benchmark resources are already useful for evaluating important subproblems, especially in remote-sensing perception and spacecraft telemetry analysis. These resources do not yet constitute a complete constellation-scale benchmark suite, but they provide an important experimental foundation for reproducible comparison.

For remote-sensing vision tasks, representative datasets include EuroSAT for land-use and land-cover classification from Sentinel-2 imagery \cite{helber2019eurosat}, BigEarthNet for large-scale multi-label Sentinel benchmarking \cite{sumbul2019bigearthnet}, SEN12MS for Sentinel-1/Sentinel-2 multimodal learning and data fusion \cite{schmitt2019sen12ms}, and So2Sat LCZ42 for global local-climate-zone classification \cite{zhu2020so2sat}. High-resolution scene understanding is further supported by AID and RESISC45 for scene classification \cite{xia2017aid,cheng2017resisc45}, PatternNet for remote-sensing image retrieval \cite{zhou2018patternnet}, fMoW for large-scale overhead imagery understanding with metadata \cite{christie2018fmow}, SpaceNet 6 for SAR--optical all-weather mapping \cite{shermeyer2020spacenet6}, and MLRSNet for multi-label high-resolution semantic scene understanding \cite{qi2020mlrsnet}. Although these datasets were not originally designed for constellation-scale collaboration, they are highly relevant for studying distributed perception, multimodal fusion, split inference, and communication-efficient feature sharing.

For spacecraft health monitoring and anomaly detection, fewer public resources are available. The OPS-SAT anomaly-detection benchmark is particularly important because it provides real in-orbit telemetry, curated anomaly labels, suggested evaluation metrics, and baseline results, making it one of the most operationally relevant resources currently available for on-orbit AI benchmarking \cite{ruszczak2025opssatad}. Earlier work on spacecraft anomaly detection with sequence models also provides a useful benchmark perspective for telemetry-driven AI methods \cite{hundman2018spacecraftanom}. More broadly, benchmark design in time-series anomaly detection deserves caution, since flawed benchmark construction can create misleading impressions of progress \cite{wu2023tsadbenchmarks}. This warning is especially relevant for Space AI, where distribution shift, missing data, and sparse labels are common rather than exceptional.

\begin{table}[H]
\caption{Representative datasets and benchmark resources relevant to distributed on-orbit Space AI.}
\label{tab:dataset_benchmarks}
\centering
\footnotesize
\resizebox{\textwidth}{!}{%
\begin{tabular}{p{3.1cm} p{3.1cm} p{4.6cm} p{5.6cm}}
\hline
\textbf{Resource} & \textbf{Modality} & \textbf{Primary task} & \textbf{Relevance to this survey} \\
\hline
EuroSAT \cite{helber2019eurosat} & Sentinel-2 optical & Land-use / land-cover classification & Lightweight EO perception benchmarks for onboard or distributed classification \\
BigEarthNet \cite{sumbul2019bigearthnet} & Sentinel-2 (and later Sentinel-1/2 variants) & Large-scale multi-label scene understanding & Useful for non-IID remote-sensing splits and federated perception studies \\
SEN12MS \cite{schmitt2019sen12ms} & Sentinel-1 + Sentinel-2 & Multimodal classification / fusion & Natural benchmark for collaborative sensing and multimodal distributed inference \\
So2Sat LCZ42 \cite{zhu2020so2sat} & Sentinel-1 + Sentinel-2 & Urban/local climate zone classification & Representative benchmark for multimodal global-scale classification under domain variation \\
AID \cite{xia2017aid} / RESISC45 \cite{cheng2017resisc45} & High-resolution optical & Scene classification & Standard remote-sensing vision benchmarks for onboard perception baselines \\
PatternNet \cite{zhou2018patternnet} & High-resolution optical & Image retrieval & Useful for communication-efficient representation learning and retrieval tasks \\
fMoW \cite{christie2018fmow} & Overhead imagery + metadata & Functional scene understanding & Relevant for metadata-aware collaborative inference and multi-source learning \\
SpaceNet 6 \cite{shermeyer2020spacenet6} & SAR + optical & All-weather building mapping & Strong benchmark for cross-modal fusion under adverse imaging conditions \\
MLRSNet \cite{qi2020mlrsnet} & High-resolution optical & Multi-label semantic scene understanding & Relevant for richer scene semantics than single-label classification \\
OPS-SAT benchmark \cite{ruszczak2025opssatad} & Real satellite telemetry & Anomaly detection & One of the few public on-orbit telemetry benchmarks with operational relevance \\
Hundman et al. \cite{hundman2018spacecraftanom} & Spacecraft telemetry & Sequence-based anomaly detection & Early benchmark reference for telemetry-driven anomaly detection methods \\
\hline
\end{tabular}%
}
\end{table}

Overall, the current benchmark landscape is richer for perception than for constellation-scale collaboration itself. This imbalance suggests an important future direction: building benchmark suites that combine sensing tasks with time-varying contact graphs, onboard resource limits, delayed labels, and fault injection, so that distributed on-orbit AI methods can be compared under more realistic mission conditions.

\section{Open Challenges and Future Directions}

Distributed on-orbit space AI is no longer limited by whether collaboration in orbit is conceptually possible; rather, the main question is which problems are mature enough for near-term progress, which ones still block deployment, and which ones require community-level coordination beyond individual algorithmic advances. To make this section more actionable, we group the open problems into three priority classes. First, some directions are \emph{technically tractable in the near term} because they can build on existing distributed optimization, edge inference, and space networking tools. Second, some challenges are \emph{blocking deployment} because they directly affect mission assurance, safety, and operational acceptance. Third, some problems are \emph{community-level priorities} whose solution depends on shared benchmarks, standards, and open ecosystems rather than on isolated methodological progress. This perspective helps distinguish research opportunities from engineering bottlenecks and coordination needs \cite{thangavel2024taso,handley2018delay,rfc4838,nistairmf2023}. Beyond nominal algorithmic performance, real mission deployment is constrained by verification and validation burden, flight-software integration, contact-plan uncertainty, and the need for conservative fallback modes when distributed collaboration becomes unavailable. These factors often determine practical feasibility as strongly as model accuracy itself.

\subsection{Near-Term Technically Tractable Directions}

\textbf{Communication-efficient collaboration.}
Among the most tractable near-term directions are methods that reduce the communication burden of distributed learning and inference. In large constellations, frequent global synchronization is rarely realistic, and practical progress is more likely to come from locality-aware collaboration, asynchronous aggregation, event-triggered communication, and utility-aware message selection \cite{handley2018delay,rfc4838,cen2024satflow,yin2025oec,hu2023etcn}. These directions are attractive because they improve performance under constraints that are already well understood---limited contact windows, message budgets, and intermittent connectivity---without requiring a full solution to the hardest certification problems.

\textbf{Cross-layer co-design of networking and AI.}
A second near-term opportunity lies in cross-layer co-design. Learning, inference, and tracking generate bursty traffic in the form of model updates, features, and tracklets, while routing and contact scheduling determine when such traffic can actually be delivered. Designing networking-for-AI policies that prioritize information value, and AI-for-networking policies that improve contact utilization and routing, is both practically useful and experimentally accessible with existing DTN abstractions and simulation tools \cite{handley2018delay,rfc4838,rfc9171,rfc9172,cen2024satflow,yin2025oec}. Compared with end-to-end autonomous decision making, these problems are often easier to evaluate quantitatively through latency, throughput, and utility-per-bit metrics.

\textbf{Benchmarkable collaborative perception and inference.}
A third tractable area is collaborative sensing and distributed inference for perception-heavy workloads, especially in vision and tracking. Split inference, early-exit models, feature sharing, and uncertainty-aware fusion can already be studied with current onboard AI pipelines and realistic networking assumptions. These problems are particularly suitable for short-term progress because they can often be benchmarked before full flight qualification, and because performance trade-offs can be reported in terms of communication, energy, and latency rather than only abstract model accuracy \cite{yin2025oec,thangavel2024taso}. In other words, these directions are research-ready because they can be iterated in simulation and hardware-in-the-loop environments without immediately facing the full burden of mission-critical certification.

\subsection{Current Blockers to Deployment}

\textbf{Verification, validation, and runtime assurance.}
The clearest blocker to deployment is not raw algorithmic capability but assurance. Learning-enabled multi-agent policies, adaptive models, and collaborative inference pipelines remain difficult to validate across the full space of operating conditions, especially when connectivity is intermittent and data distributions drift over time. For deployment, the key requirement is not only good nominal performance, but defensible evidence that unsafe or unstable behaviors will be detected and bounded. This makes runtime assurance, regression testing, fault injection, staged rollout, and conservative fallback policies central bottlenecks for constellation-scale autonomy \cite{cofer2020rta,hobbs2023rta,thangavel2024taso,nistairmf2023}. In practice, these issues often determine whether a system can be allowed to close a control or operations loop at all.

\textbf{Fault tolerance and silent corruption.}
A second deployment blocker is robustness to the space environment. Radiation-induced faults, transient resets, silent data corruption, and degraded sensors can invalidate both distributed learning and distributed inference, even if the communication network itself is functioning correctly \cite{oliveira2016gpu,wang2023mars,souvatzoglou2024seu}. While robust aggregation and conservative fusion can mitigate some failures, truly dependable deployment requires fault-aware update validation, uncertainty-aware collaboration, and graceful degradation policies that revert to local-only or conservative modes when trust signals deteriorate. This is a harder problem than ordinary packet-loss robustness, because the system must reason about when received information is merely delayed and when it is unsafe.

\textbf{Safe model lifecycle management.}
Model updates in orbit are another major blocker. Continual adaptation is attractive under drift and non-IID data, but operationally it creates risk: a poorly validated update may degrade performance in a way that is hard to recover from during the mission. For this reason, constellation-scale Space AI needs explicit lifecycle controls, including versioning, signing, validation gates, staged deployment, rollback, and continuous monitoring after release \cite{thangavel2024taso,nistssdf2022,slsa2024}. This is less a pure machine learning problem than an engineering-governance problem. Until update governance becomes routine, many learning-enabled capabilities will remain limited to advisory roles rather than mission-authoritative ones.

\textbf{Security and adversarial resilience.}
Security is also deployment-blocking in settings where nodes may be compromised or links may be contested. Poisoning, backdoors, and malicious track/model updates are particularly serious in distributed settings because local corruption can propagate through aggregation and fusion \cite{bagdasaryan2020backdoor,blanchard2017krum}. Existing DTN and CCSDS security mechanisms such as BPSec provide an important foundation for integrity and confidentiality, but integrating these protections with learning-enabled workflows, update pipelines, and collaborative decision systems remains underdeveloped \cite{rfc9172,ccsds7345bpsec,nasa1006a,bailey2021spacecyber}. From a deployment perspective, this makes security not a secondary enhancement but part of the core safety case.

\subsection{Community-Level Coordination Priorities}

\textbf{Benchmarks, datasets, and reproducible digital twins.}
Some of the most important gaps cannot be closed by a single research group. Chief among them is the lack of shared evaluation substrates that reflect constellation reality: dynamic contact graphs, non-IID data, concept drift, delayed labels, energy budgets, and fault injection. The field needs benchmarks and digital twins that combine orbit propagation, communications, onboard compute, and AI workloads in a reproducible way. Existing open tools such as Basilisk, GMAT, and Orekit provide strong building blocks, but they must be connected to common tasks, shared data splits, and standardized reporting protocols to become true community benchmarks \cite{kenneally2020basilisk,hughes2014gmatvv,orekit2024zenodo}. This is a coordination problem as much as a technical one.

\textbf{Standards and interoperable interfaces.}
Another community priority is standardization. Distributed on-orbit AI will remain difficult to compare and hard to integrate unless researchers converge on practical interfaces for exchanging model updates, feature tensors, uncertainty-aware tracklets, and mission metadata. CCSDS and DTN standards already define realistic communication and security baselines, but they do not yet provide a mature common layer for learning-enabled collaboration \cite{ccsds133spacepacket,ccsds727cfdp,rfc9171,rfc9172,ccsds7342bp,ccsds7345bpsec}. Progress here requires coordination among AI researchers, flight-software developers, mission operators, and standards bodies, rather than algorithmic innovation alone.

\textbf{Open ecosystems and transparent reporting.}
Finally, the field needs open-source ecosystems and transparent documentation practices. Curated repositories of papers, code, datasets, and reproducible baselines can reduce fragmentation and accelerate research-to-deployment transfer. Similarly, documentation frameworks such as datasheets and model cards are important because they make assumptions, limitations, and intended use explicit, which is especially valuable in safety- and mission-critical settings \cite{gebru2021datasheets,mitchell2019modelcards}. In this sense, open ecosystems are not merely a convenience; they are part of the infrastructure required for comparability, accountability, and community growth.

\begin{table}[H]
\caption{Priority-based roadmap for constellation-scale distributed on-orbit AI.}
\label{tab:open_roadmap}
\centering
\footnotesize
\resizebox{\textwidth}{!}{%
\begin{tabular}{p{3.2cm} p{2.8cm} p{5.8cm} p{3.8cm} p{4.8cm}}
\hline
\textbf{Direction} & \textbf{Priority class} & \textbf{Why it matters now} & \textbf{Main actors} & \textbf{Suggested evaluation signals} \\
\hline
Communication-efficient collaboration & Near-term tractable & Directly improves feasibility under limited contacts and message budgets; can be tested with existing simulators and prototypes & Individual research groups + mission AI teams & Bytes/contact, staleness, utility per bit, energy/latency trade-off \\
\hline
Cross-layer AI/network co-design & Near-term tractable & Networking already constrains collaboration; utility-aware scheduling can deliver immediate system gains & AI + networking researchers & End-to-end latency, throughput, delivery timeliness, uncertainty reduction per bit \\
\hline
Collaborative perception and distributed inference & Near-term tractable & Supports practical EO/SSA workloads and can be benchmarked before full autonomy certification & AI/vision groups + onboard computing teams & Accuracy vs latency/energy, early-exit rate, feature-sharing cost, robustness under drops \\
\hline
Verification, validation, and runtime assurance & Deployment blocker & Determines whether learning-enabled autonomy can be trusted in safety- or mission-critical loops & Flight software, assurance, autonomy teams & Constraint violations avoided, intervention rate, regression coverage, fault-injection robustness \\
\hline
Fault tolerance and silent-corruption resilience & Deployment blocker & Radiation and transient failures can invalidate distributed updates and fusion results & Hardware/software reliability + autonomy teams & Recovery time, safe fallback rate, consistency under injected faults \\
\hline
Safe model lifecycle management & Deployment blocker & In-orbit updates require governance, rollback, and monitoring before operational adoption & Mission operators + MLOps/flight software teams & Update success/failure rate, rollback triggers, post-update regression signals \\
\hline
Security and adversarial resilience & Deployment blocker & Compromised nodes or links can undermine learning, fusion, and coordination at system scale & Cybersecurity + autonomy teams & Attack success rate, poisoned-update robustness, integrity-check effectiveness \\
\hline
Benchmarks and digital twins & Community coordination & No shared basis yet exists for fair comparison under constellation realities & Research community + agencies + industry & Public tasks, common splits, fault/contact models, reproducibility of reported results \\
\hline
Standards and interoperable interfaces & Community coordination & Needed for cross-mission reuse, comparability, and integration with space networking stacks & Standards bodies + agencies + industry & Protocol compliance, interoperability demonstrations, reusable schemas/interfaces \\
\hline
Open ecosystems and transparent reporting & Community coordination & Reduces fragmentation and supports accountable, cumulative progress & Entire community & Availability of code/data, documentation quality, reuse across studies \\
\hline
\end{tabular}%
}
\end{table}

Overall, the most tractable near-term opportunities lie in communication-efficient collaboration, cross-layer AI/network co-design, and benchmarkable collaborative perception. By contrast, the main blockers to real deployment are assurance, fault tolerance, security, and safe update governance, because these determine whether learning-enabled autonomy can be trusted operationally. Finally, shared benchmarks, interoperable standards, and open ecosystems are community-level priorities that no single paper or laboratory can solve alone. For constellation-scale on-orbit AI to mature beyond promising prototypes, progress on all three levels will be necessary, but the order matters: practical short-term advances should be pursued in parallel with assurance-driven engineering, while the community collectively builds the common infrastructure needed for long-term field maturity.

\section{Conclusion \& Discussion}

This survey reviewed distributed on-orbit space AI for satellite constellations, focusing on how multiple satellites can collaborate to learn, infer, and act under realistic space constraints. We introduced a constellation-centric system model that captures dynamic connectivity, heterogeneous onboard resources, and non-IID, drifting data streams, and proposed a taxonomy that organizes methods by collaboration architecture, timing mechanism, and trust assumptions.

We synthesized the literature through three complementary algorithmic pillars. Federated learning enables cross-satellite training and adaptation without centralizing raw data, but demands contact-aware protocols that tolerate intermittent participation, limit communication overhead, and support safe, auditable update workflows. Multi-agent coordination addresses constellation-level autonomy for coverage, scheduling, formation, and collision avoidance, where partial observability and delayed communication are intrinsic and hard constraints must be enforced throughout planning and execution. Collaborative sensing and distributed inference improve constellation-level perception via multi-satellite fusion, tracking, and partitioned inference, and underscore the importance of cross-layer co-design with networking and compute offloading.

We further summarized representative applications in Earth observation, space situational awareness, autonomous constellation networking, and on-orbit servicing, and emphasized evaluation practices that jointly measure task performance and system-level costs using end-to-end simulation and reproducible benchmarks. Looking ahead, the main barriers to deployment are scalability to mega-constellations, trustworthiness under faults and adversaries, and operational integration across the mission lifecycle. Progress will depend on methods that remain effective under limited and intermittent communication, degrade gracefully under uncertainty and failures, and provide conservative governance for in-orbit model updates. We hope this review clarifies the design space and supports the development of practical, verifiable, and scalable approaches to constellation-scale on-orbit space AI .

\bibliography{sample}

\end{document}